\ifdefined\pdfminorversion
\fi
\ifdefined\pdfsuppresswarningpagegroup
\fi

\documentclass[10pt,twocolumn,letterpaper]{article}
\usepackage[pagenumbers]{cvpr}

\makeatletter
\def\abstract{%
   \iftoggle{cvprpagenumbers}{}{\thispagestyle{empty}}%
   \centerline{\large\bf Abstract}%
   \vspace*{0.4em}\noindent%
   \it\ignorespaces%
}
\makeatother

\def\paperID{}
\def\confName{CVPR}
\def\confYear{2026}

\usepackage{color}
\usepackage{epsfig}
\usepackage{graphicx}

\usepackage{booktabs}  %
\usepackage{tabularx}               %
\newcolumntype{C}{>{\centering\arraybackslash}X}
\usepackage{multirow}               %
\usepackage{diagbox}                %
\usepackage{hhline}                 %
\usepackage{color}                  %

\usepackage{booktabs}
\usepackage{extarrows}
\usepackage{makecell}
\usepackage{wrapfig}
\usepackage{colortbl}
\usepackage{xcolor}
\usepackage{longtable}

\usepackage{amssymb}

\usepackage{adjustbox}
\usepackage{array}
\usepackage{floatflt}
\usepackage{stfloats} %
\usepackage{placeins} %

\usepackage{bm}
\usepackage{nicefrac}
\usepackage{microtype}

\usepackage{changepage}
\usepackage{extramarks}
\usepackage{fancyhdr}
\usepackage{lastpage}
\usepackage{setspace}
\usepackage{soul}
\usepackage{xspace}

\usepackage[pagebackref=true,breaklinks=true,colorlinks,citecolor=citecolor,anchorcolor=citecolor,linkcolor=citecolor,urlcolor=citecolor,bookmarks=false]{hyperref}
\usepackage{url}

\usepackage{enumerate}
\usepackage{enumitem}  %

\usepackage{makecell}

\usepackage{pifont} %

\usepackage{algpseudocode}
\usepackage{xcolor}

\usepackage[symbol]{footmisc}

\usepackage{caption}

\usepackage{etoc} %

\usepackage{scalefnt}

\usepackage{fontawesome5}

\usepackage{siunitx}

\usepackage{listings}

\usepackage[ruled,vlined]{algorithm2e} %
\usepackage{wrapfig,float}

\newcolumntype{L}[1]{>{\raggedright\let\newline\\\arraybackslash\hspace{0pt}}m{#1}}
\newcolumntype{R}[1]{>{\raggedleft\let\newline\\\arraybackslash\hspace{0pt}}m{#1}}

\newcommand{\ignore}[1]{}

\makeatletter
\DeclareRobustCommand\onedot{\futurelet\@let@token\@onedot}
\def\@onedot{\ifx\@let@token.\else.\null\fi\xspace}

\makeatother

\definecolor{MyBlue}{rgb}{0.46, 0.50, 0.61}
\definecolor{MyDarkBlue}{rgb}{0,0.08,0.8}
\definecolor{MyDarkGreen}{RGB}{45,155,45}
\definecolor{MyDarkRed}{rgb}{0.8,0.02,0.02}
\definecolor{MyOrange}{rgb}{1.0, 0.4, 0.2}
\definecolor{MyPurple}{RGB}{111,0,255}
\definecolor{MyRed}{rgb}{0.8,0.0,0.0}
\definecolor{MyGold}{rgb}{0.75,0.6,0.12}
\definecolor{MyDarkgray}{rgb}{0.66, 0.66, 0.66}
\definecolor{MyBrown}{rgb}{0.65, 0.16, 0.16}
\definecolor{MyMutedRose}{rgb}{0.58, 0.29, 0.35}
\definecolor{JiayuanColor}{rgb}{0.60,0.43,0.48}
\definecolor{erranColor}{rgb}{24, 40, 113}
\definecolor{WenlongColor}{RGB}{0,120,215}
\definecolor{YuWeiColor}{RGB}{118,185,0}

\definecolor{citecolor}{HTML}{696FAD}

\newcommand{\name}{$\textsc{MindCube}$\xspace}

\newif\ifpropositionfirstitem
\propositionfirstitemtrue

\definecolor{bggray}{HTML}{F5F5F5}
\definecolor{pvdblue}{HTML}{DAE8FC}
\definecolor{RoseQuartzBg}{HTML}{F7CAC9}
\definecolor{RoseQuartz}{HTML}{F5A798}
\definecolor{Serenity}{HTML}{92A8D1}
\definecolor{OrangeRed}{rgb}{1.0, 0.27, 0.0}
\definecolor{RoyalBlue}{cmyk}{1, 0.50, 0, 0}
\definecolor{Turquoise}{HTML}{0F4C81}
\definecolor{mint}{rgb}{0.24, 0.71, 0.54}
\definecolor{green}{rgb}{0.0, 0.120, 0.0}

\newdimen\abovecrulesep
\newdimen\belowcrulesep
\makeatletter
\patchcmd{\@@@cmidrule}{\aboverulesep}{\abovecrulesep}{}{}
\patchcmd{\@xcmidrule}{\belowrulesep}{\belowcrulesep}{}{}
\makeatother

\definecolor{mybluetitle}{HTML}{4B527E} %

\definecolor{mygreen}{RGB}{0,150,0}
\definecolor{boxbackground}{HTML}{F0F7FF}  %
\definecolor{boxborder}{HTML}{D0D9E5}      %
\definecolor{accentblue}{HTML}{4A86E8}     %
\definecolor{lightblue}{HTML}{EEF3FF}  %
\definecolor{bordergray}{HTML}{CCCCCC}  %
\definecolor{headerblue}{HTML}{2C5AA0}  %

\definecolor{lavenderframe}{HTML}{E6E6FA}  %
\definecolor{lighterlav}{HTML}{F5F5FF}  %
\definecolor{codegray}{rgb}{0.5,0.5,0.5}  %
\definecolor{codepurple}{HTML}{483D8B}  %
\definecolor{backcolour}{HTML}{F5F5FF}  %
\lstdefinestyle{mystyle}{
    backgroundcolor=\color{backcolour},
    commentstyle=\color{headerblue},
    keywordstyle=\color{codepurple},
    numberstyle=\tiny\color{codegray},
    stringstyle=\color{codepurple},
    basicstyle=\ttfamily\scriptsize,
    breakatwhitespace=false,
    breaklines=true,
    captionpos=b,
    keepspaces=true,
    frame=none,
    numbersep=5pt,
    showspaces=false,
    showstringspaces=false,
    showtabs=false,
    tabsize=2
}

\definecolor{jsonkey}{RGB}{44, 130, 201}     %
\definecolor{jsonstring}{RGB}{255, 140, 0}   %
\definecolor{jsonnumber}{RGB}{34, 139, 34}   %

\lstdefinelanguage{json}{
    basicstyle=\ttfamily\small,
    numbers=left,
    numberstyle=\tiny\color{gray},
    stepnumber=1,
    numbersep=5pt,
    showstringspaces=false,
    breaklines=true,
    frame=none,
    backgroundcolor=\color{gray!5},
    literate=
     *{:}{{{\color{jsonkey}:}}}{1}
      {,}{{{\color{jsonkey},}}}{1}
      {"}{{{\color{jsonstring}"}}}{1}
      {[}{{{\color{jsonkey}[}}}{1}
      {]}{{{\color{jsonkey}]}}}{1}
      {0}{{{\color{jsonnumber}0}}}{1}
      {1}{{{\color{jsonnumber}1}}}{1}
      {2}{{{\color{jsonnumber}2}}}{1}
      {3}{{{\color{jsonnumber}3}}}{1}
      {4}{{{\color{jsonnumber}4}}}{1}
      {5}{{{\color{jsonnumber}5}}}{1}
      {6}{{{\color{jsonnumber}6}}}{1}
      {7}{{{\color{jsonnumber}7}}}{1}
      {8}{{{\color{jsonnumber}8}}}{1}
      {9}{{{\color{jsonnumber}9}}}{1}
}

\colorlet{osfirst}{teal!50}
\colorlet{ossecond}{teal!30}
\colorlet{osthird}{teal!10}
\colorlet{lavenderfirst}{violet!50}
\colorlet{lavendersecond}{violet!30}
\colorlet{lavenderthird}{violet!10}

\usepackage{amsmath,amsfonts,bm}

\def\eqref#1{equation~\ref{#1}}

\def\1{\bm{1}}

\DeclareMathAlphabet{\mathsfit}{\encodingdefault}{\sfdefault}{m}{sl}
\SetMathAlphabet{\mathsfit}{bold}{\encodingdefault}{\sfdefault}{bx}{n}

\DeclareMathOperator*{\argmin}{arg\,min}

\newif\ifpwTightParagraphSpacing
\pwTightParagraphSpacingfalse

\makeatletter
\ifpwTightParagraphSpacing
  \def\pwParagraphBeforeSkip{1.0ex\@plus0.3ex\@minus0.2ex}
  \def\pwParagraphAfterSkip{-0.8em}
\else
  \def\pwParagraphBeforeSkip{3.25ex\@plus1ex\@minus.2ex}
  \def\pwParagraphAfterSkip{-1em}
\fi

\renewcommand\paragraph{\@startsection{paragraph}{4}{\z@}%
  {\pwParagraphBeforeSkip}%
  {\pwParagraphAfterSkip}%
  {\normalfont\normalsize\bfseries}}
\makeatother

\DeclareRobustCommand{\name}{\textup{\textls[-15]{\textsc{PointWorld}}}\xspace}
\DeclareRobustCommand{\algo}{\name}
\title{PointWorld: Scaling 3D World Models for In-The-Wild Robotic Manipulation}

\author{
\textbf{Wenlong Huang$^{1,\dagger}$,
Yu-Wei Chao$^{2}$,
Arsalan Mousavian$^{2}$,
Ming-Yu Liu$^{2}$}\\
\textbf{Dieter Fox$^{2}$,
Kaichun Mo$^{2,*}$,
Li Fei-Fei$^{1,*}$}\\[1ex]
$^{1}$Stanford University\quad
$^{2}$NVIDIA\\
$^{*}$\,Equal advising\quad
$^{\dagger}$\,Work done partly at NVIDIA
}

\begin{document}

\twocolumn[{%
\renewcommand\twocolumn[1][]{#1}%
\maketitle
\vspace{-2.8em}
\begin{center}
    {\normalsize \href{https://point-world.github.io}{\textcolor{magenta}{point-world.github.io}}}
\end{center}
\vspace{-1.em}
\begin{center}
    \centering
    \captionsetup{type=figure}
    \includegraphics[width=\textwidth]{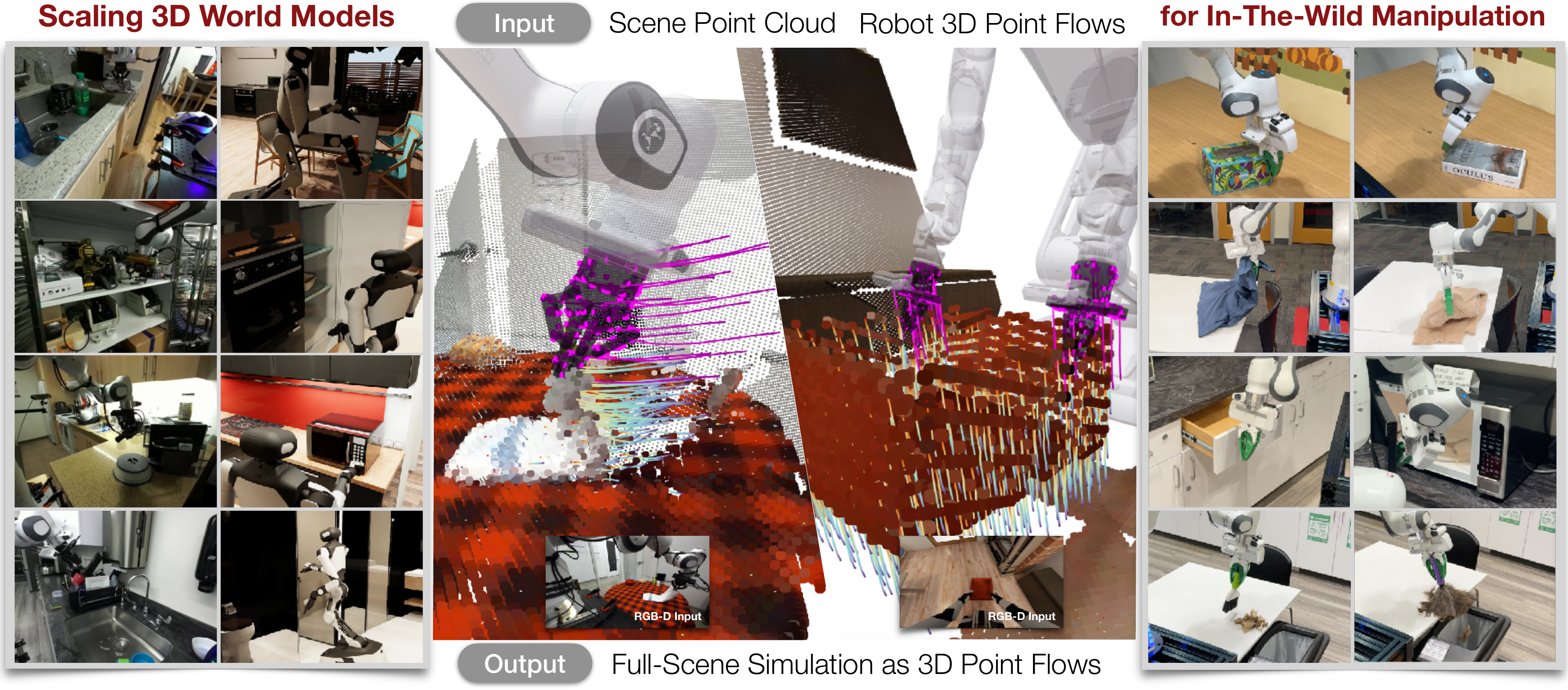}
    \vspace{-1.6em}
    \caption{\textbf{\algo} is a large pre-trained 3D world model that predicts full-scene 3D point flows from a static point cloud and an embodiment-agnostic description of robot actions, represented also as 3D point flows.
    We curate a large-scale 3D dynamics modeling dataset, spanning single-arm, bimanual, whole-body, mobile manipulation interactions in real and simulated domains.
    Through careful evaluations, we rigorously study the recipe for scaling up 3D world models.
    Pretrained on diverse data, a single model enables diverse manipulation behaviors on physical hardware, given only a single RGB-D image captured in the wild, without additional data or finetuning.
    }
    \label{fig:teaser}
\end{center}
}]

\begin{abstract}
Humans anticipate, from a glance and a contemplated action of their bodies, how the 3D world will respond, a capability that is equally vital for robotic manipulation.
We introduce \textbf{\algo}, a large pre-trained 3D world model that unifies state and action in a shared 3D space as 3D point flows: given one or few RGB-D images and a sequence of low-level robot action commands, \algo forecasts per-pixel displacements in 3D that respond to the given actions.
By representing actions as 3D point flows instead of embodiment-specific action spaces (e.g., joint positions), this formulation directly conditions on physical geometries of robots,
while seamlessly integrating learning across embodiments.
To train our 3D world model, we curate a large-scale dataset spanning real and simulated robotic manipulation in open-world environments, enabled by recent advances in 3D vision and simulated environments, totaling about 2M trajectories and 500 hours across a single-arm Franka and a bimanual humanoid.
Through rigorous, large-scale empirical studies of backbones, action representations, learning objectives, partial observability, data mixtures, domain transfers, and scaling, we distill design principles for large-scale 3D world modeling.
With a real-time (0.1s) inference speed, \algo can be efficiently integrated in the model-predictive control (MPC) framework for manipulation.
We demonstrate that a \textbf{single pre-trained checkpoint} enables a real-world Franka robot to perform rigid-body pushing, deformable and articulated object manipulation, and tool use, without requiring any demonstrations or post-training and all from a single image captured in-the-wild.
Code, dataset, and pre-trained checkpoints will be open-sourced.
\end{abstract}

\vspace{-2em}
\section{Introduction}

\begin{figure*}[t]
    \centering
    \includegraphics[width=\textwidth]{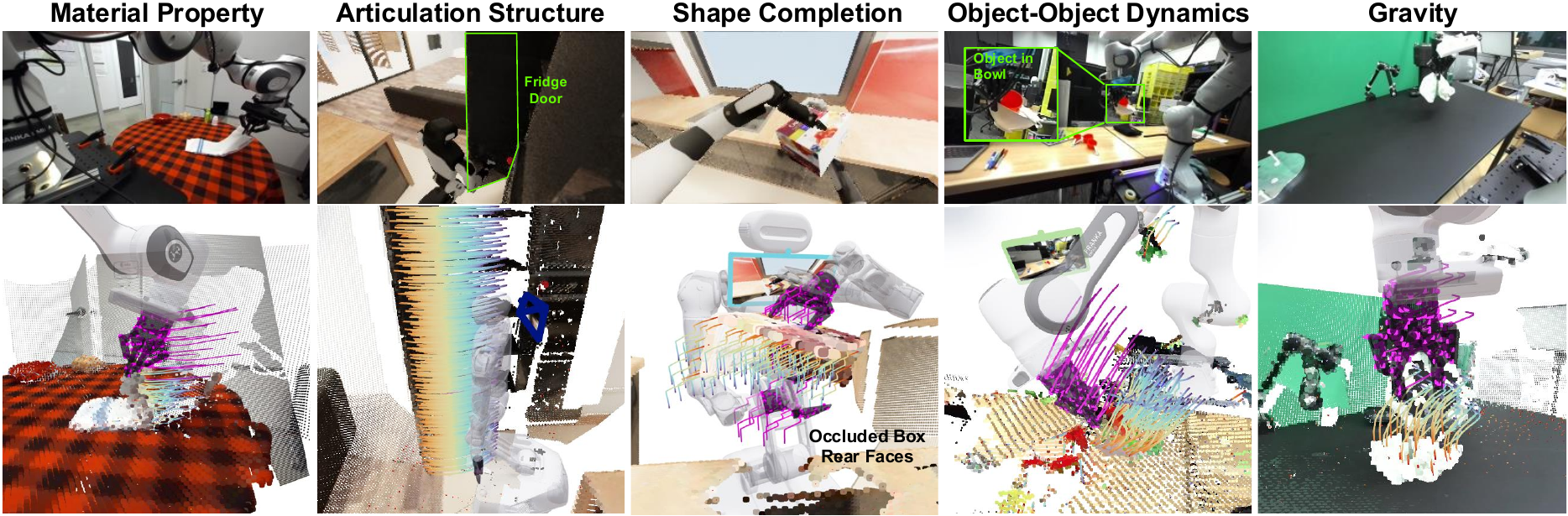}
    \vspace{-1.9em}
    \caption{\textbf{Rich Supervision of 3D World Modeling for Physical Interactions}, when conditioned on \textcolor{magenta}{3D robot point flows} and partial observable RGB-D.
    The 3D world modeling objective enjoys dense pixel-level supervision while encoding a wide range of capabilities central to robotic manipulation.
    To predict full-scene evolution, the model needs to implicitly segment objects of interest, identify material property and/or articulation structure, perform implicit shape completion for contact reasoning, propagate robot-object interaction for object-object dynamics, and simultaneously considering the effects of gravity, encapsulated all in a single forward pass of the learned model.
    }
    \label{fig:challenging}
    \vspace{-1.5em}
\end{figure*}

World modeling in unstructured environments is imperative for general-purpose robots: predicting how the world evolves from what the robot sees and intends to do with its body. Humans do this from a glance and a grasp, forecasting deformation, articulation, stability, and contact, revealing how much a world-modeling objective captures when conditioned on a contemplated action in 3D (Figure~\ref{fig:challenging}). Actions unfold where physics lives, in space and time: our aim is a predictive model that makes such spatially grounded, action-conditioned predictions from only perceptual inputs in open-world settings, a pinnacle goal of spatial intelligence~\cite{li2025fromwordstoworlds}.

A large body of work has studied world modeling from complementary angles. Physics-based models~\citep{mujoco2012}, while capable of highly accurate predictions, face sim-to-real gaps and require curated, environment-specific modeling.
Learning-based dynamics models~\citep{li2018learning} address this by learning from observed interaction, yet often depend on domain-specific inductive bias (e.g., full observability, objectness priors, or material specification).
In parallel, large video generative models trained at scale~\citep{videoworldsimulators2024} are capable of producing photorealistic predictions but lack explicit action conditioning and often fall short on physical consistency.
See \citet{ai2025_survey} for a recent survey.
Despite progress, a gap remains between what current models predict and what humans can foresee from visual observations in the wild and a contemplated action.

Our philosophy is unification for scaling: represent \emph{state} and \emph{action} in the same modality of 3D physical space. State is represented by a full-scene 3D point cloud built from RGB-D captures; actions are dense 3D point trajectories instantiated from the agent's own embodiment, typically known a priori (e.g., a robot description file), and thus forecastable over time. Under this representation, 3D world modeling equates to modeling \emph{full-scene 3D point flow} under perturbations from a temporal sequence of robot points: given partially observed 3D scene points and those action points, predict per-point scene displacements over a horizon.
While conceptually simple, this formulation ties raw sensory observation and an \emph{embodiment-agnostic} action space in a shared representation through dynamics (what moves, how, and where) and implicitly captures objectness, articulation, and material properties, all through interaction between the robot's specific geometry (e.g., grippers, fingers) and the partially-observed scene.
By modeling the \emph{geometries of interaction} independent of goals, \algo aims to capture the single source of truth of the physical world, while naturally learning from heterogeneous embodiments, tasks, and trajectories (regardless of success or failure), akin to ``next-token prediction''~\citep{gpt} but for interaction over 3D space and time.
We term our approach \textbf{\algo}.

To provide supervision, we curate a large-scale dataset for 3D dynamics modeling, spanning hundreds of in-the-wild scenes with single-arm, bimanual, and whole-body interactions across both real and simulated domains. The dataset was built from existing robotic manipulation datasets, DROID~\citep{droid} and BEHAVIOR-1K~\citep{behavior}.
Since accurate 3D annotations are crucial for capturing precise contact in physical interactions,
significant efforts were spent to build a custom pipeline to extract 3D point flows from the real-world dataset, enabled by recent advances in metric depth estimation~\citep{foundationstereo}, camera pose estimation~\citep{vggt}, and point tracking~\citep{karaev2025cotracker3}.
Leveraging the dataset, we distill important design decisions for large-scale 3D dynamics learning through rigorous investigations of backbone architectures, action representations, objectives, partial observability, data mixtures, scaling laws, and domain transfers under zero-shot and finetuned settings.

To demonstrate \algo's potential for manipulation, we integrate it with a model-predictive controller (MPC) for action inference on a real robot. As \algo predicts scene dynamics jointly over short action chunks in a single forward pass at a real-time latency (0.1s), it provides a natural and efficient integration with sampling-based MPC (e.g., MPPI~\citep{mppi}).
We show that a \textbf{single pre-trained checkpoint} enables a real-world robot to perform rigid-body pushing, deformable and articulated object manipulation, and tool use, without requiring any demonstrations or post-training and all from a single image captured in-the-wild.
 
\noindent \textbf{Contributions.}
\textbf{(i)} We introduce a large pre-trained 3D world model, \algo, that unifies state and action in a shared representation of 3D point flows, and present rigorous studies of its modeling recipe.
\textbf{(ii)} We curate and open-source a large-scale high-quality 3D interaction dataset used for training \algo, totaling \(\sim2\)M trajectories or \(\sim500\) hours.
\textbf{(iii)} We demonstrate a single pre-trained \algo enables a real robot to perform diverse manipulation tasks from a single in-the-wild RGB-D capture, without requiring additional demonstrations or training.

\begin{figure}[t]
    \centering
    \includegraphics[width=\linewidth]{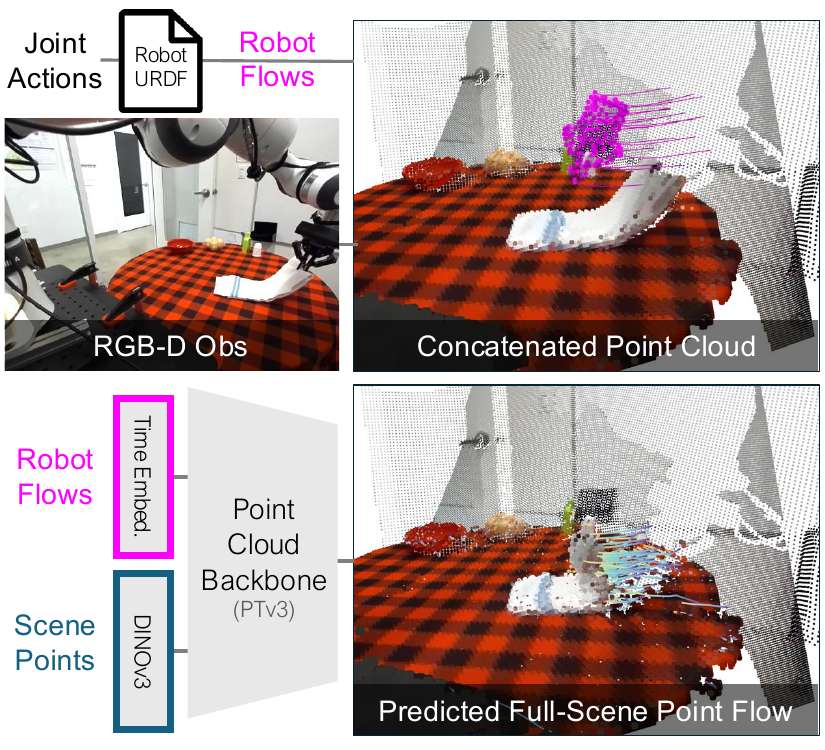}
    \vspace{-2.3em}
    \caption{\textbf{Overview of \algo}. Given calibrated RGB-D, robot joint-space actions, and a robot description file (URDF), we convert actions to robot flows and concatenate with scene to form a single point cloud serving as an \emph{embodiment-agnostic interaction geometry}. Scene points are featurized with a frozen DINOv3 encoder, robot points with temporal embeddings, and a point cloud backbone predicts \emph{full-scene} 3D point flows.}
    \label{fig:method}
    \vspace{-1.7em}
\end{figure}

\section{Related Work}

\paragraph{World Modeling.}
World models~\citep{world_models_ha} are predictive models that simulate future states given current state and action, categorized often by their state-action representations.
Video models use pixel-space state, trained either with photometric reconstruction~\citep{finn2016unsupervised,pixel_pred_savp,video_action_cond_1,prednet2016,svg2018,videogpt2021,video_diffusion_1,makeavideo2022,imagenvideo2022,animatediff2023,dynamicrafter2023,videocrafter22023,videoworldsimulators2024,videopoet_2023,pvgdm_2023,zhen2025tesseract,kotar2025world} or joint-embedding predictions~\citep{vjepa_2024,vjepa2_2025}.
3D world models instead operate on meshes or explicit surfaces~\citep{neus,meshgraphnets,edonet_2022,phystwin2025,xia2025drawer,xia2024video2game,physgen3d2025}, radiance fields or Gaussians~\citep{nerf,gaussian_splatting,driess2023learning,pegs_2024,real_is_sim_2025,physdreamer2024,gwm2025,zhang2025real,physgaussian2024,li2025wonderplay}, or particles~\citep{interaction_networks,gns,li2018learning,particlenerf_2022,particleformer_2025,zhang2025particle,zhang2024adaptigraph,he2025scaling,whitney2024modeling}.
Hybrid approaches additionally reason over hierarchical structures in world modeling~\citep{aeronautiques1998pddl,kaelbling2011hierarchical,mao2019neuro,lecun2022path,critiques_world_models,liu2024world,wang2025enact}.
Action parameterizations range from low-level joint-space commands~\citep{finn2016unsupervised,planet,dreamer,dreamerv2,tdmpc2,du2025_gpc_world_modeling,guo2025ctrl,li2025robotic}, to camera and navigation motions~\citep{bruce2024genie,zhou2025stable,cameractrl,matrixgame2,hunyuan_gamecraft,cami2v,song2025generative,zhou2025learning}, textual prompts~\citep{unisim_2023,videoworldsimulators2024,cosmos_2025,cosmos_transfer1,flare2025,dreamgen2025}, and 2D cues~\citep{go_with_the_flow,mofa_video,motion_prompting,force_prompting,animate_anyone,motion_i2v,motionctrl,draganything,dragnuwa,tora,tooncrafter}.
Robot actions can then be produced by online planning~\citep{cem2005,mppi,kelly2017_collocation,pineau2003_pbvi,planet,li2018learning}, offline policy synthesis~\citep{planet,dreamer,dreamerv2,mbpo,pets2018,hafner2025training,tdmpc2}, or inverse-dynamics models~\citep{agrawal2016learning,du2023learning,act_from_actionless_2023}.
\algo uses 3D point flow as shared state-action representation, emphasizing contact and geometry rather than appearance,
conditions on 3D actions with specific geometry of given robot/gripper,
interaction beyond only visible regions compared to 2D cues,
and doing so with one (or sparse) input images (with estimated depth) in a single, real-time forward pass of a large pre-trained model (Figure~\ref{fig:challenging}).

\paragraph{Dynamics Models in Robotics.}\label{sec:related:dynamics}
Dynamics models in robotics instantiate world models with robot action spaces. They include physics-based simulators~\citep{mujoco2012,bullet2015,drake,isaacgym2021,gazebo2004,difftaichi2020} and learning-based models~\citep{ai2025_survey,interaction_networks,gns,meshgraphnets,li2018learning,particlenerf_2022,particleformer_2025,zhang2025particle,zhang2024adaptigraph,guo2025ctrl,wu2023daydreamer}.
Crucial for robotics, they support policy learning~\cite{peng2018sim,andrychowicz2020learning,kumar2021rma}, planning~\cite{lozano1983spatial,kaelbling1998planning,kaelbling2011hierarchical,toussaint2015logic,marcucci2024shortest}, model-based RL~\citep{planet,dreamer,dreamerv2,mbpo,pets2018,hafner2025training,tdmpc2}, exploration and online guidance~\citep{pathak2017curiosity,sekar2020planning,mordatch2014combining,du2025dynaguide}, safety filtering~\citep{ames2019control,sun2025latent}, model-based design and verification~\cite{donze2010breach,drake,xu2024dynamics}, and policy evaluation~\citep{barreiros2025careful,zhang2025real,wpe_2025,team2025evaluating}.
While existing dynamics models often require curated, scene-specific modeling~\cite{ai2025_survey}, our aim is to pre-train a single dynamics model that generalizes across diverse in-the-wild environments.
Using 3D flows as state-action space, it naturally encapsulates many action parameterizations used in prior works in an embodiment-agnostic manner: joint-space commands~\cite{wu2023daydreamer,guo2025ctrl,wang2025precise}, end-effector actions~\cite{zhang2025particle}, and motion primitives~\cite{li2018learning,chen2023predicting}, while operating on partially observable RGB-D image(s) in the wild without scene reconstruction~\cite{phystwin2025}, priors on objectness~\cite{zhang2025particle} or materials~\cite{mpm_sulsky1993}.

\paragraph{2D and 3D Flows for Manipulation.}
Flows (or point tracks), which address correspondences across space and time, provide a powerful interface between perception and control.
With advances in point tracking~\cite{karaev2025cotracker3,doersch2024bootstap,xiao2024spatialtracker}, recent works explored them as structured representations for policy learning~\cite{wen2023any,haldar2025point,weng2021fabricflownet,seita2023toolflownet,wang2025skil,guo2025flowdreamer,duisterhof2024deformgs,yin2025object}, reward modeling~\cite{xu2024flow,patel2025real,guzey2025bridging,shipoints2reward}, (sub-)goal specification~\cite{huang2024rekep}, or as visual servoing targets~\cite{eisner2022flowbot3d,act_from_actionless_2023,bharadhwaj2024track2act,patel2025rigvid,li2025novaflow,dream2flow}.
In this work, we leverage recent advances in 3D vision (depth~\citep{foundationstereo}, camera pose estimation~\citep{vggt}, and point tracking~\citep{karaev2025cotracker3}) to label 3D scene flows from large-scale real-world manipulation dataset~\cite{droid} (with robot flows obtained from known robot geometry, kinematics, and proprioception), which enables training of a large 3D world models via stable regression losses to capture robotic interactions with diverse objects in open-world environments.

\section{Method}
\label{sec:method}

We formulate 3D world modeling as action-conditioned full-scene 3D point flow prediction (Section~\ref{sec:method:algo}; Figure~\ref{fig:method}). We then describe how \algo may be used for action inference and discuss its use case in the framework of model predictive control that we explore in this work (Section~\ref{sec:method:robotics}).

\subsection{3D World Modeling with \textbf{\algo}}
\label{sec:method:algo}
We model environment dynamics as a neural network
$\mathcal{F}_\theta : \mathbf{S} \times \mathbf{A} \rightarrow \mathbf{S}$
parametrized by $\theta$
that predicts next state given current state and robot action, where $\mathbf{S}$ and $\mathbf{A}$ denote state and action spaces.
Existing approaches~\cite{ai2025_survey} typically formulate this as a single-step update
$\mathbf{s}_{t+1} \,=\, \mathcal{F}_\theta\big(\mathbf{s}_t, \mathbf{a}_t\big)$.
In contrast, we adopt a multi-step (chunked) formulation for data-driven modeling~\cite{zhao2023learning}: the model predicts future states over a horizon $H$ in a single forward pass
\(\mathcal{F}^{H}_\theta : (\mathbf{s}_t, \mathbf{a}_{t:t+H-1}) \rightarrow \mathbf{s}_{t+1:t+H}\),
which improves temporal consistency and amortizes computation.
We use $H=10$ steps and 0.1s per step.

\paragraph{State Representation.}
Building a world model requires a deliberate choice of a state space $\mathbf{S}$, with state at time $t$ denoted by $\mathbf{s}_t \in \mathbf{S}$.
In this work, we use \emph{point flows} (referred also as particles~\cite{li2018learning,zhang2025particle}) as the environment state.
Formally, let $\mathbf{s}_t = \{\,(\mathbf{p}_{t,i},\, \mathbf{f}^S_i)\,\}_{i=1}^{N_S}$ denote the point flows at time $t$, consisting of $N_S$ points with positions $\mathbf{p}_{t,i}\in\mathbb{R}^3$ and time-constant features $\mathbf{f}^S_i\in\mathbb{R}^{D_S}$ of dimension $D_S$ for each point.
Compared to alternative representations, point flows offer the following advantages for world modeling in manipulation:
(i) emphasis on physical interactions between 3D geometries instead of appearance, akin to the role of physics simulators rather than renderers;
(ii) accessibility from any RGB-D captures in partially observable environments~\cite{kaelbling1998planning} while not assuming objectness or material priors;
(iii) simple and stable training via L2 losses on displacements, without permutation matching;
(iv) expressiveness to capture diverse fine-grained contact dynamics.
To obtain the point flows, from one or a few calibrated RGB‑D views, we mask robot pixels via forward kinematics (using the URDF and joint configuration) and back‑project the remaining pixels to obtain $\mathbf{p}_{t,i}$.
Note that since the model takes in a static point set from the environment as input, and correspondence is preserved only within the model's forward pass (i.e., its ``imagination''), no separate point tracker is required for inference, and point count may vary between forward passes.

\paragraph{Action Representation.}
To learn from heterogeneous embodiments (different kinematics, gripper geometries, and even different numbers of grippers),
we again use 3D point flows.
However, unlike scene point flows which are obtained from RGB-D captures, robot point flows are generated by forecasting the robot’s own geometry via forward kinematics using its URDF (known a priori).
This is an intentional design for ensuring ``imagined actions'' are \emph{fully}, rather than partially, observable while being represented in an \emph{embodiment-agnostic} way–crucial in cases where contact occurs in occluded regions (e.g., holding and transporting a large box with egocentric view).
Specifically, given a sequence of joint configurations $\{\mathbf{q}_{t+k}\}_{k=0}^H$, we sample robot surface points once at time $t$, attach each to its corresponding link, and propagate them with forward kinematics to obtain an ordered set of $N_R$ robot points $\{\,(\mathbf{r}_{t+k,j},\,\mathbf{f}^R_{t+k,j})\,\}_{j=1}^{N_R}$ at each time step $t{+}k$,
where $\mathbf{r}_{t+k,j}\in\mathbb{R}^3$ denotes the position of point $j$ at time $t{+}k$ and $\mathbf{f}^R_{t+k,j}\in\mathbb{R}^{D_R}$ is its time-varying feature vector of dimension $D_R$.
We treat this collection as the action at time $t{+}k$ and denote it by $\mathbf{a}_{t+k}$.
This yields an embodiment‑agnostic description of \emph{interaction geometry} over the horizon.
In practice, most robot surface points never contact the scene; for efficiency, we sample robot point flows from only the grippers (a few hundred points per gripper depending on its geometry). See Section~\ref{sec:exp:action} for experiments.

\begin{figure*}[t]
    \centering
    \includegraphics[width=\textwidth]{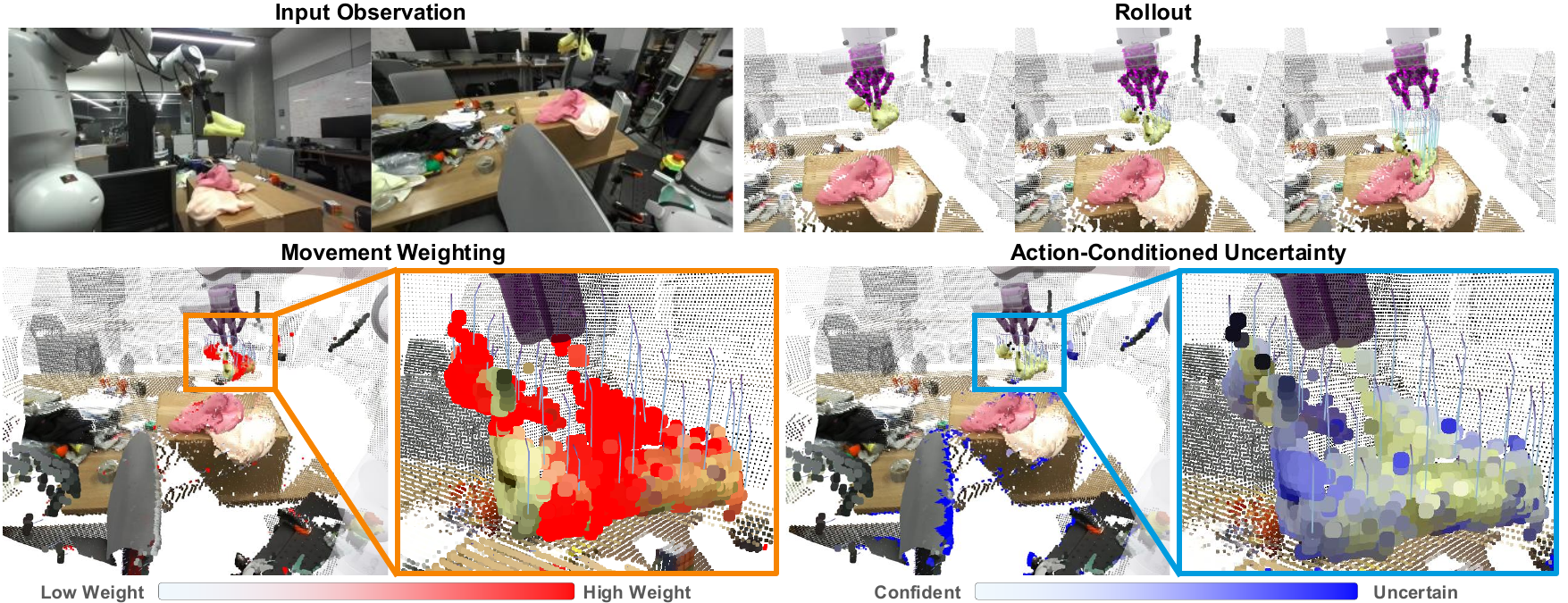}
    \vspace{-1.9em}
    \caption{\textbf{Movement Weighting and Uncertainty Regularization}, where the robot releases and drops a yellow cloth.
    (\textbf{Bottom Left}) The movement weighting, used in the training objective, effectively biases the training towards scene points that are moving at each timestep, computed with the ground-truth flows.
    (\textbf{Bottom Right}) The uncertainty value, predicted by the model without any ground-truth, regularizes training to prevent overfitting to points that have unreliable ground-truth.
    Intriguingly, we observe that it also emerges to capture action-conditioned uncertainty arising from the object's physical properties (e.g., larger variability along the edge of the cloth).
    }
    \label{fig:movement-uncertainty}
    \vspace{-1.6em}
\end{figure*}

\paragraph{Dynamics Prediction.}
Given the above state-action representations, we now have a static full-scene point cloud $\mathbf{s}_t$ and a temporal sequence of robot point-flow actions $\mathbf{a}_{t:t+H-1}$ as inputs to the model.
Instead of designing custom architectures, we deliberately build on top of state-of-the-art point cloud backbones~\cite{ptv3} to distill the core principles that enable scalable, large-scale 3D world modeling.
Towards this goal, we concatenate the initial scene points with the time‑stacked robot points to form a single point cloud processed by the backbone.
Scene points are featurized with frozen DINOv3~\citep{dinov3,rayst3r} by projecting them to 2D views, while robot points are featurized with temporal embedding.
The point cloud backbone processes the concatenated point cloud and outputs features for all points.
A shared MLP head then predicts per-point displacements of the scene points at each step within a chunk of length $H$ in a single forward pass.
This chunked formulation delivers extremely efficient inference capable of evaluating many candidate trajectories with a real-time latency (\(0.1\,\mathrm{s}\) per batched forward pass), which stands in contrast to pixel-based approaches that typically require seconds-long inference~\cite{vjepa2_2025,cosmos_2025} due to the use of diffusion objectives.

\paragraph{Training Objective.}
While the formulation lends itself to standard regression objectives, 3D world modeling introduces two distinctive challenges that require careful design:
(i) due to full-scene prediction, the robot often manipulates only a small subset of the scene, so most points are static and standard L2 loss leads to very sparse training signal;
(ii) real-world data is noisy, so we need to regularize the model to be robust to this noise.
To address challenge (i), we adopt a weighted regression objective, reweighting each point at each timestep by a soft movement likelihood $m_{k,i}\in[0,1]$ computed from ground-truth motion so as to focus the loss on moving points.
Letting $\delta_{k,i}\ge 0$ denote the norm of the ground-truth displacement vector for point $i$ at step $k$, we set $m_{k,i} = \sigma\big(\kappa(\delta_{k,i} - \tau)\big)$, where $\sigma$ is the logistic sigmoid, and $\tau$ and $\kappa$ are non-negative displacement-threshold and temperature parameters, respectively.
We then normalize these likelihoods to obtain weights $w_{k,i} = m_{k,i} / \sum_{k,i} m_{k,i}$ for each point $i$ at step $k$.
To address challenge (ii), we adopt aleatoric uncertainty regularization~\cite{kendall2017uncertainties,novotny2017learning,vggt} by predicting a scalar log-variance $s_{k,i}$ for each point $i$ at step $k$ and further using a Huber loss on the residual.
Formally, the full training objective becomes:
\vspace{-0.5em}
\begin{equation}
\tfrac{1}{2}\sum_{\substack{k, i}}^{\substack{H, N_S}}
\underbrace{w_{k,i}}_{\substack{\text{\scriptsize movement}\\\text{\scriptsize weight}}}
\Big(
\underbrace{\rho_\delta\big(\mathbf{\hat P}_{t+k,i} - \mathbf{P}_{t+k,i}\big)}_{\substack{\text{\scriptsize Huber loss}\\\text{\scriptsize on 3D residual}}}
\underbrace{e^{-s_{k,i}}}_{\substack{\text{\scriptsize uncertainty}\\\text{\scriptsize weight}}}
+
\underbrace{s_{k,i}}_{\substack{\text{\scriptsize uncertainty}\\\text{\scriptsize reg.}}}
\Big),
\end{equation}
where $\rho_\delta$ is the elementwise Huber loss, and $\mathbf{\hat P}_{t+k,i}$ and $\mathbf{P}_{t+k,i}$ are the predicted and ground-truth positions of point $i$ at step $k$, respectively.
In practice, we also ignore the points that are deemed not visible by the 2D tracker used to provide the pseudo ground-truth (more details in Section~\ref{sec:dataset}).

\subsection{\textbf{\algo} for Robotic Manipulation}
\label{sec:method:robotics}
A pre-trained \algo enables diverse use cases in robotics, as discussed in Section~\ref{sec:related:dynamics}.
In this work, we specifically investigate whether a single pre-trained \algo can enable action inference in unseen, in-the-wild real-world environments from only a single RGB-D capture, \textbf{without any additional demonstrations or post-training at deployment time}.
To this end, we integrate \algo in an MPC framework with a sampling-based planner MPPI~\citep{mppi} that plans a sequence of $T$ end‑effector pose targets in $\mathrm{SE}(3)$ given a cost function defined in the model's state space.
Specifically, given a calibrated RGB-D capture, we first form a scene point set as described in Section~\ref{sec:method:algo}, yielding an initial state $\mathbf{s}_0$.
We then sample $K$ action perturbations $\ell_{1:K}$ using a time-correlated (cubic-spline) noise distribution, which are added to a nominal end‑effector trajectory.
For each sampled trajectory $\mathbf{E}^{(\ell)}_{1:T}$, the corresponding robot point-flow actions $\mathbf{a}_{1:T}^{(\ell)}$ are constructed, scene flows are rolled out by \algo conditioned on $\mathbf{a}_{1:T}^{(\ell)}$, and a trajectory cost $J^{(\ell)}$ is accumulated.
The nominal trajectory is iteratively refined by computing exponentiated weights
\(
\omega_\ell \propto \exp(-J^{(\ell)}/\beta)
\)
over samples and updating the nominal as a weighted average of sampled trajectories, where $\beta$ is non-negative temperature.

To define the cost function, we separate task objectives from control regularization.
Let $\mathcal{I}_{\text{task}}\subseteq\{1,\dots,N_S\}$ denote a set of task-relevant scene points, with associated target positions $\{\mathbf{g}_i\}_{i\in\mathcal{I}_{\text{task}}}$. Task cost on a predicted state $\mathbf{s}_k$ at time $k$ is
\(
c_{\text{task}}(\mathbf{s}_k) = \tfrac{1}{|\mathcal{I}_{\text{task}}|}\sum_{i\in\mathcal{I}_{\text{task}}} \|\mathbf{p}_{k,i}-\mathbf{g}_i\|_2^2.
\)
Such pointwise goal costs apply broadly across rigid, deformable, and articulated objects.
Task-relevant points can be specified by either human via GUI or by VLMs~\cite{huang2024rekep}.
The overall optimization problem is formulated as a global trajectory optimization:
\begin{equation}
\label{eq:mpc}
\begin{aligned}
\argmin_{\mathbf{E}_{0:T}}\; & \sum_{k=1}^T \Big[\, c_{\text{task}}\big(\mathbf{s}_k\big) \;+\; c_{\text{ctrl}}\big(\mathbf{E}_k\big) \,\Big] \\
\text{s.t.}\; & \mathbf{s}_{1:T} \,=\, \mathcal{F}^{T}_\theta\!\big(\,\mathbf{s}_0,\; \mathbf{a}_{1:T}\,\big), \; \mathbf{E}_0 \,=\, \mathbf{E}_{\text{measured}},
\end{aligned}
\end{equation}
where $c_{\text{ctrl}}$ subsumes path length and reachability regularization,
$\mathbf{E}_k$ denotes end‑effector pose at step $k$,
and $\mathbf{E}_{\text{measured}}$ is the current end‑effector pose. Further details in Appendix.

\section{Dataset Curation and Evaluation Protocol}
\label{sec:dataset}

\begin{figure*}[t]
	\centering
	\includegraphics[width=0.95\linewidth]{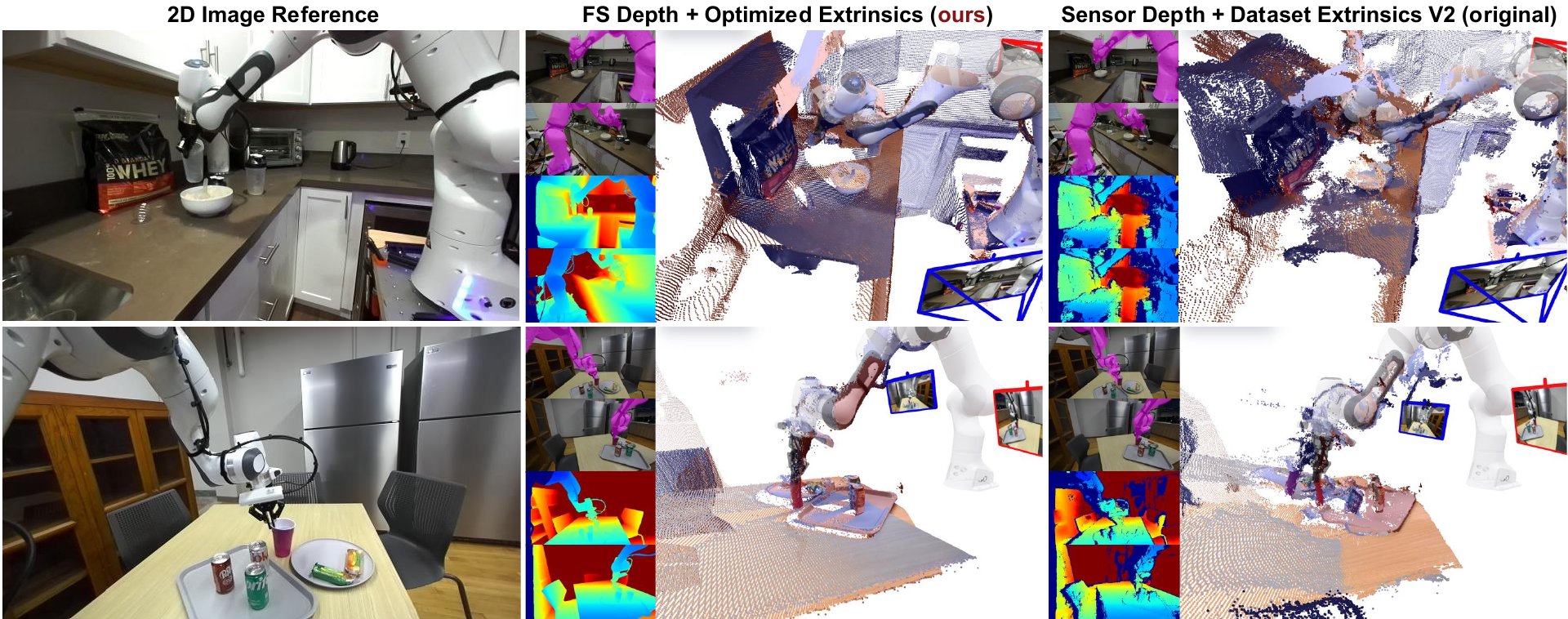}\\[0.4em]
	\includegraphics[width=0.95\linewidth]{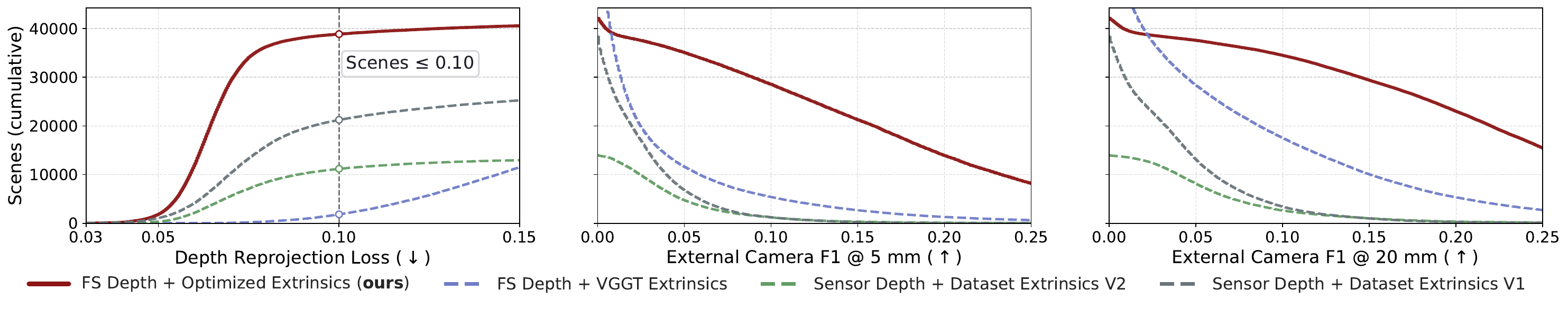}
	\vspace{-1.1em}
	\caption{\textbf{3D Annotation Quality and Comparisons.}
	FS denotes FoundationStereo~\citep{foundationstereo}; Dataset Extrinsics V1 and V2 are the two DROID extrinsics releases. 
	\textbf{(Top)} Compared to DROID releases, our pipeline yields substantially higher-quality depth and camera pose calibration, resulting in more accurate robot mask overlays and better aligned point clouds (readers are encouraged to zoom in or check out the interactive visualization on the \href{https://point-world.github.io}{project website} for details).
	\textbf{(Bottom)} We further compute depth reprojection loss (differences between analytical and observed depth of robot surface), and F1 scores of point cloud alignment.
	We observe purely leveraging existing models (FS, VGGT) are insufficient, and V2 extrinsics improve over V1 by filtering out scenes with poor point cloud alignment but result in significantly lower scene counts.
	In contrast, our annotation pipeline retains substantially more scenes below \(0.10\) depth-loss criterion and dominates all metrics.}
	\label{fig:annotation}
	\vspace{-0.7em}
\end{figure*}

\begin{figure*}[!h]
    \centering
    \includegraphics[width=0.99\textwidth]{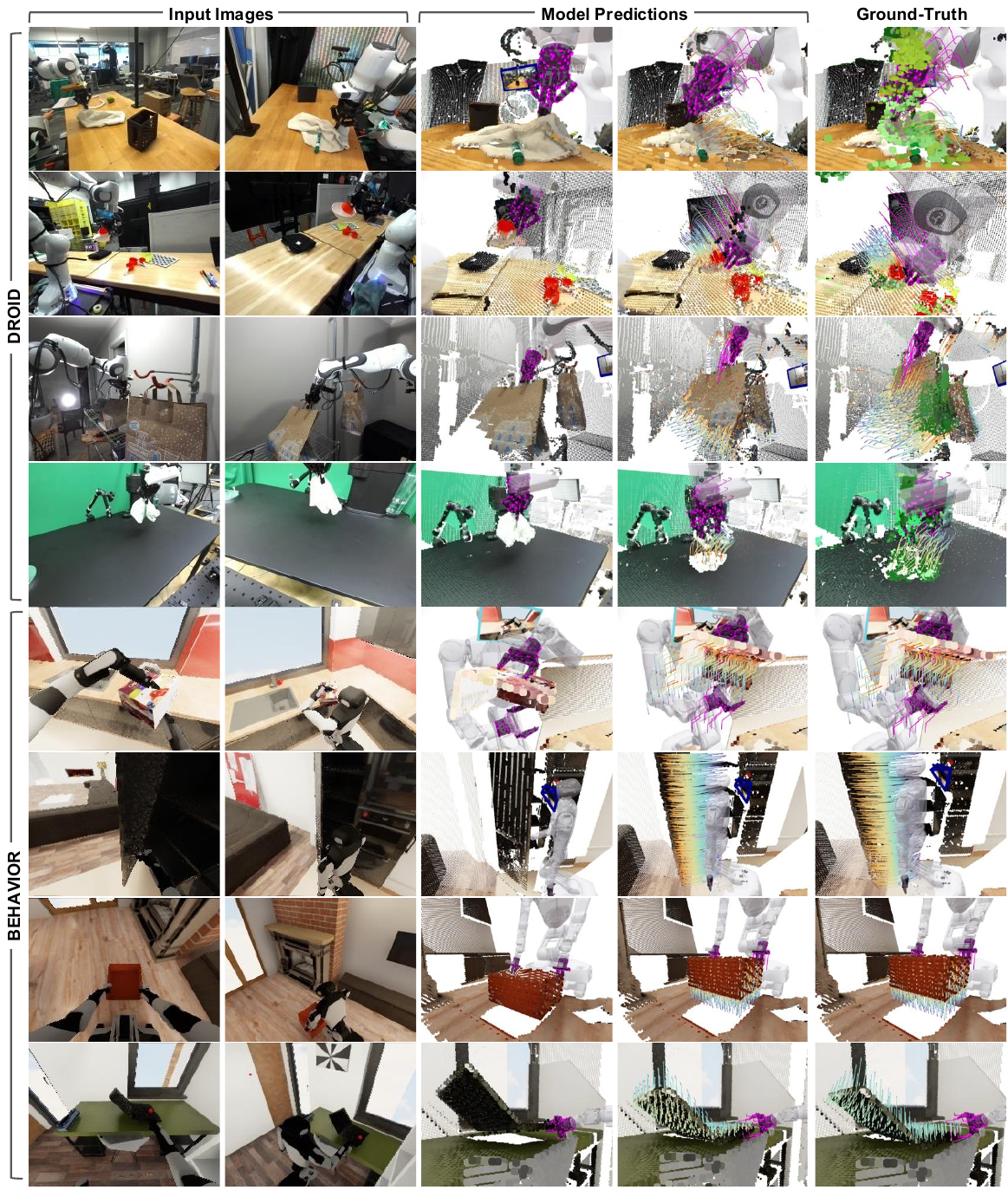}
	\vspace{-0.8em}
    \caption{\textbf{Unseen rollouts from a single pre-trained \algo across diverse domains}, visualized with Viser~\cite{yi2025viser}. Given RGB-D captures, \algo predicts 10-step point flows conditioned on \textcolor{magenta}{robot flows}. We show first prediction, last prediction, and last ground-truth. Green points in GT mark regions occluded during 2D point tracking, for which we observe model predictions are often more accurate because these points are not being supervised in model training.
    Due to grid downsampling ($1.5$cm) we apply to all point clouds, all 3D visualization is upsampled to image resolution for visual clarity via nearest neighbors.
	Interactive visualization at \href{https://point-world.github.io}{project website}.
	}
    \label{fig:rollouts}
\end{figure*}

\begin{figure}[t]
	\centering
	\includegraphics[width=0.9\linewidth]{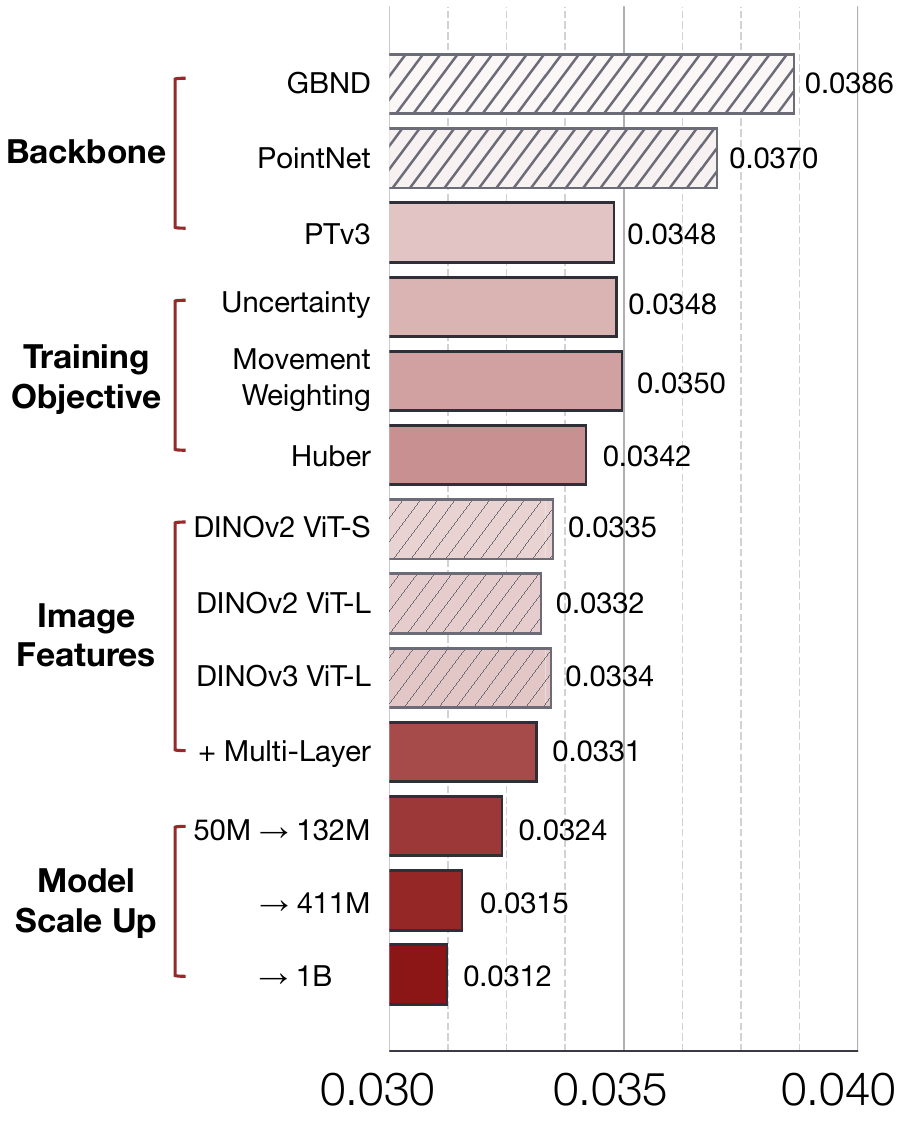}
	\vspace{-0.85em}
	\caption{\textbf{Roadmap for Scaling 3D World Models}, measured by $\ell_2$ error on moving scene points on DROID test set. Starting from an existing baseline~\cite{ai2025_survey}, we progressively modernize the backbone, stabilize training objectives, leverage pre-trained features, and scale model size, yielding consistent gains in accuracy. Hatched bars indicate settings that are not adopted in the final model.}
	\label{fig:progress}
	\vspace{-1.5em}
\end{figure}
	
	Accurate, large-scale 3D data is essential for the world model in Section~\ref{sec:method} to generalize in the wild.
	Apart from requiring action labels, the dataset needs to also have accurate spatial perception (i.e., high-fidelity depth), hand-eye calibration (i.e., camera extrinsics in robot base frame), and per-pixel correspondence matching amid occlusions (i.e., point tracking).
	While large efforts have been made for collecting diverse real-world manipulation datasets~\cite{droid,o2024open}, obtaining their 3D annotations has previously been challenging.
	Our key observation is that recent advances in 3D vision—metric depth estimation, camera pose estimation, and dense point tracking—are maturing to provide a markerless offline pipeline that operates purely on recorded data to produce such a dataset of interest (\figurename~\ref{fig:annotation} top-left).
	Photorealistic simulation complements this with ground-truth supervision.
	Combining both, we curate a dataset of about \(2\)M trajectories (\(500\) hours) spanning single-arm, bimanual, and whole-body teleoperated interactions across in-the-wild real scenes and simulated home-scale environments.
	To the best of our knowledge, this constitutes the largest 3D dynamics modeling dataset, which we fully open-source.

	\paragraph{3D Annotation for Real-World Data.}
	We leverage DROID~\cite{droid}, a robot manipulation dataset with diverse in-the-wild interactions recorded by two external cameras and a wrist-mounted camera.
	Although DROID provides sensor depth and camera extrinsics, the depth often degrades in open-world environments and camera poses are inaccurate due to imperfect calibration.
	Frontier 3D reconstruction models such as VGGT~\citep{vggt} jointly estimate depth and camera parameters from RGB images and often look visually plausible, but yield overly smoothed depth maps and camera poses that can deviate from ground-truth by tens of centimeters.
	
	After extensive experimentation, we adopt a three-stage annotation pipeline that combines several learned models with a dedicated optimization procedure.
	First, we replace sensor depth with stereo-estimated depth from FoundationStereo~\citep{foundationstereo}, which is particularly effective at the close working distances typical of manipulation.
	Second, we compute camera extrinsics by refining VGGT-initialized camera poses with an optimization procedure that aligns robot depth observations to the known robot mesh.
	Third, given accurate depth and extrinsics, we perform per-pixel point tracking using CoTracker3~\citep{karaev2025cotracker3}.
	CoTracker3 is a 2D point tracker that outputs image-space correspondences and a visibility map; we lift these tracks to 3D using the refined depth and camera poses and carry over the visibility labels so that occluded points are excluded from supervision during model training.
	With this pipeline, we recover reliable tracked 3D point flows for over 60\% of DROID (nearly 200 hours of raw human teleoperation) and obtain reconstructed point clouds that both qualitatively and quantitatively improve over both original dataset and alternative annotation methods (\figurename~\ref{fig:annotation}).
	To further assess extrinsics accuracy in the absence of ground truth in the real world, we treat the best 1\% of scenes under the original dataset extrinsics (as measured by depth reprojection loss) as a proxy for the true extrinsics.
	Relative to this reference, our optimized extrinsics achieve a median translation and rotation error of 1.8~cm and 1.9~degrees.
	
	\paragraph{Simulation (BEHAVIOR-1K).}
	To complement real-world data, we use BEHAVIOR-1K~\citep{behavior} (B1K), which provides about \(1100\) hours of teleoperated (pre-filtering) interaction in photorealistic home-scale environments with bimanual, whole-body, and mobile manipulation.
	We obtain ground-truth 3D point flows by leveraging known simulation state.
	Because the dataset focuses on long-horizon activities while \algo focuses on short-horizon interaction dynamics, we filter trajectories using privileged information accessible in simulation.
	We retain only trajectories with active contacts between robot and objects and those with nonzero object motion.
	More details are in Appendix.
	
	\paragraph{Model Evaluation Protocol.}
	We evaluate predicted point flow from \algo and other baselines using a per-point, per-timestep $\ell_2$ distance over the prediction horizon.
	Because most scene points remain static during robot interaction, we focus on the metric on moving points ($\ell_2$ mover), as measured by ground-truth data and filter the full set of points using the movement likelihood introduced in Section~\ref{sec:method}.
	For real-world domains, we further denoise the evaluation data by training a separate expert model only on the held-out test split to flag unreliable flows via the uncertainty objective from Section~\ref{sec:method}, retaining only the top 80\% of points measured by model confidence.
	All evaluated models are trained exclusively on the imperfect training set and are evaluated on the expert-filtered test set.
	Details in Appendix.
	
	\paragraph{Interpretation of the Metric.}
	This dense per-point $\ell_2$ metric is highly discriminative when comparing methods and reveals systematic differences in rollout fidelity that task-level success rates often fail to expose~\cite{barreiros2025careful}.
	Because all errors are measured over one-second horizons, absolute metric differences can appear modest, since even large motions move points by only a few centimeters, yet we empirically observe that small numerical differences often correspond to pronounced qualitative gains in rollout fidelity.
	Given the scale of the evaluation set (approx. $40{,}000$ robot trajectories with $10{,}000$ point flows each), standard errors for the $\ell_2$ metrics are negligible ($\leq 10^{-5}$m), so we report means only.

\section{Experiments}
\label{sec:experiments}

Focusing on real-world data, we chart a roadmap of empirical lessons we learned for scaling 3D world models (Section~\ref{sec:exp:roadmap}, \figurename~\ref{fig:progress})~\cite{liu2022convnet,sonata}.
We then discuss targeted ablations along complementary design axes for \algo (Section~\ref{sec:exp:ablations}).
Using real and simulated data, we quantify in-domain, cross-domain, and held-out generalization under zero-shot and finetuned settings (Section~\ref{sec:exp:generalization}).
Finally, we study \algo for MPC-based action inference on a physical robot in the wild without extra demonstrations or finetuning (Section~\ref{sec:exp:robotics}).
All experiments are constructed to isolate a single modeling choice under controlled setups compared to baselines unless otherwise stated.

\begin{table}[t]
    \centering
    \definecolor{backboneStaticCol}{HTML}{EDEDED}
    \definecolor{backboneDot}{HTML}{8B1D1D}
    \newcommand{\basebackbone}{\textcolor{backboneDot}{\ding{109}}\,}
    \newcommand{\bestbackbone}{\textcolor{backboneDot}{\ding{108}}\,}
    {\footnotesize
    \setlength{\tabcolsep}{3pt}
    \renewcommand{\arraystretch}{1.05}
    \begin{adjustbox}{max width=0.98\linewidth}
    \begin{tabular}{@{}l|r>{\columncolor{backboneStaticCol}}r>{\columncolor{backboneStaticCol}}r>{\columncolor{backboneStaticCol}}r|r>{\columncolor{backboneStaticCol}\arraybackslash}r@{\hspace{2pt}}}
        \toprule
        \textbf{Backbone} & \textbf{Params} & \textbf{Mem.} & \textbf{FLOPs} & \textbf{Latency} & \textbf{$\ell_2$ mov.} & \textbf{$\ell_2$ stat.} \\
        \midrule
        \basebackbone GBND & $1.00\mathrm{x}$ & $1.00\mathrm{x}$ & $1.00\mathrm{x}$ & $13.46$ & 0.0390 & 0.0066 \\
        \basebackbone PointNet & $1.03\mathrm{x}$ & $0.34\mathrm{x}$ & $0.04\mathrm{x}$ & $5.93$ & 0.0369 & 0.0084 \\
        \basebackbone PointNet++ & $1.07\mathrm{x}$ & $0.67\mathrm{x}$ & $0.06\mathrm{x}$ & $327.08$ & 0.0368 & 0.0073 \\
        \basebackbone SparseConv & $33.31\mathrm{x}$ & $7.18\mathrm{x}$ & $1.32\mathrm{x}$ & $17.70$ & 0.0396 & 0.0076 \\
        \basebackbone Transformer & $41.06\mathrm{x}$ & $0.31\mathrm{x}$ & $3.38\mathrm{x}$ & $30.43$ & 0.0339 & 0.0071 \\
        \bestbackbone PTv3-50M & $49.14\mathrm{x}$ & $0.30\mathrm{x}$ & $0.34\mathrm{x}$ & $59.60$ & 0.0331 & 0.0067 \\
        \bestbackbone PTv3-132M & $127.22\mathrm{x}$ & $0.69\mathrm{x}$ & $1.04\mathrm{x}$ & $69.60$ & 0.0324 & 0.0061 \\
        \bestbackbone PTv3-411M & $398.67\mathrm{x}$ & $1.89\mathrm{x}$ & $1.90\mathrm{x}$ & $102.47$ & 0.0315 & 0.0059 \\
        \bestbackbone PTv3-1B & $\mathbf{957.71\mathrm{x}}$ & $4.30\mathrm{x}$ & $3.57\mathrm{x}$ & $123.65$ & \textbf{0.0312} & \textbf{0.0056} \\
        \bottomrule
    \end{tabular}
    \end{adjustbox}}
    \vspace{-0.7em}
\caption{\textbf{Backbone Comparisons.} PTv3~\cite{ptv3} enables massive parameter scaling while retaining similar memory and efficient inference (latency in milliseconds, PTv3-1B $\approx 0.12$\,s).}
\label{tab:backbone}
\vspace{-1em}
\end{table}

\begin{figure*}[t]
    \vspace{-0.8em}
    \centering
    \captionsetup{aboveskip=2pt, belowskip=2pt}
    \begin{minipage}[c]{0.61\textwidth}
        \centering
        \includegraphics[width=\linewidth]{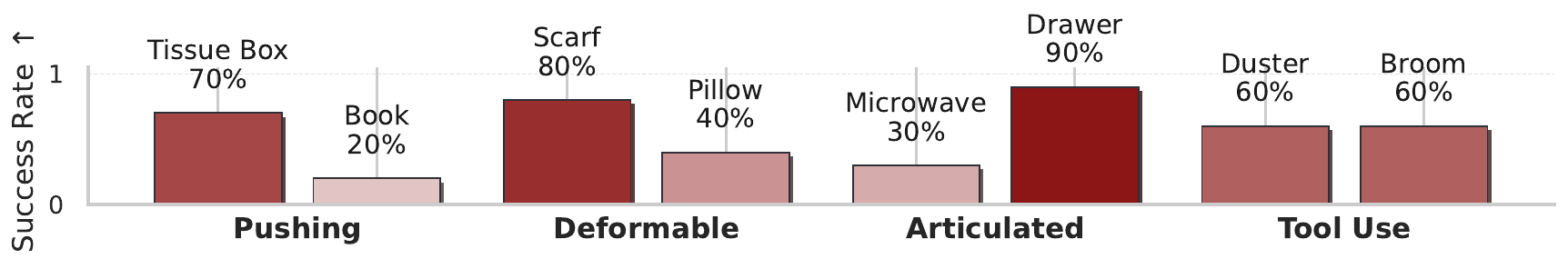}\\[0.20em]
        \includegraphics[width=\linewidth]{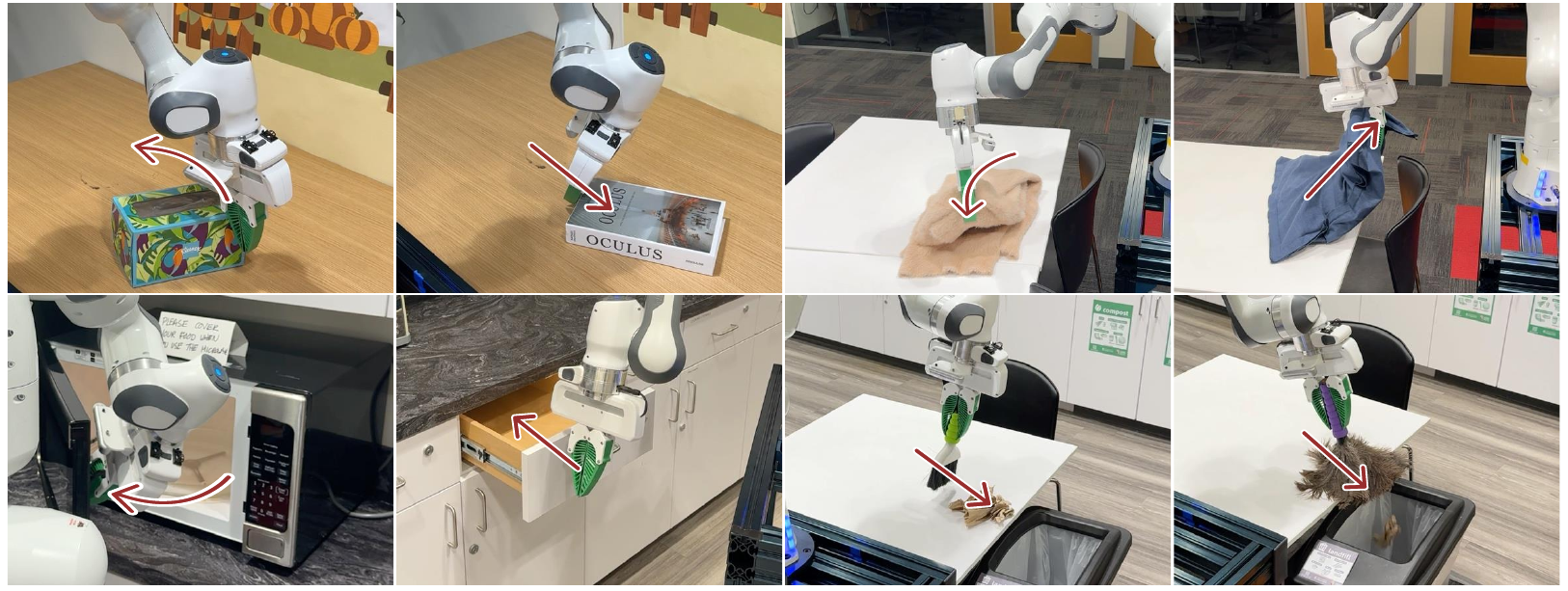}
        \captionof{figure}{\textbf{Real-World Action Inference}. \algo runs zero-shot with MPC for rigid, deformable, articulated, and tool-use tasks in the wild. Success rates are on top.}
        \label{fig:robot-success}
    \end{minipage}
    \hfill
    \begin{minipage}[c]{0.37\textwidth}
        \centering
        \includegraphics[width=\linewidth]{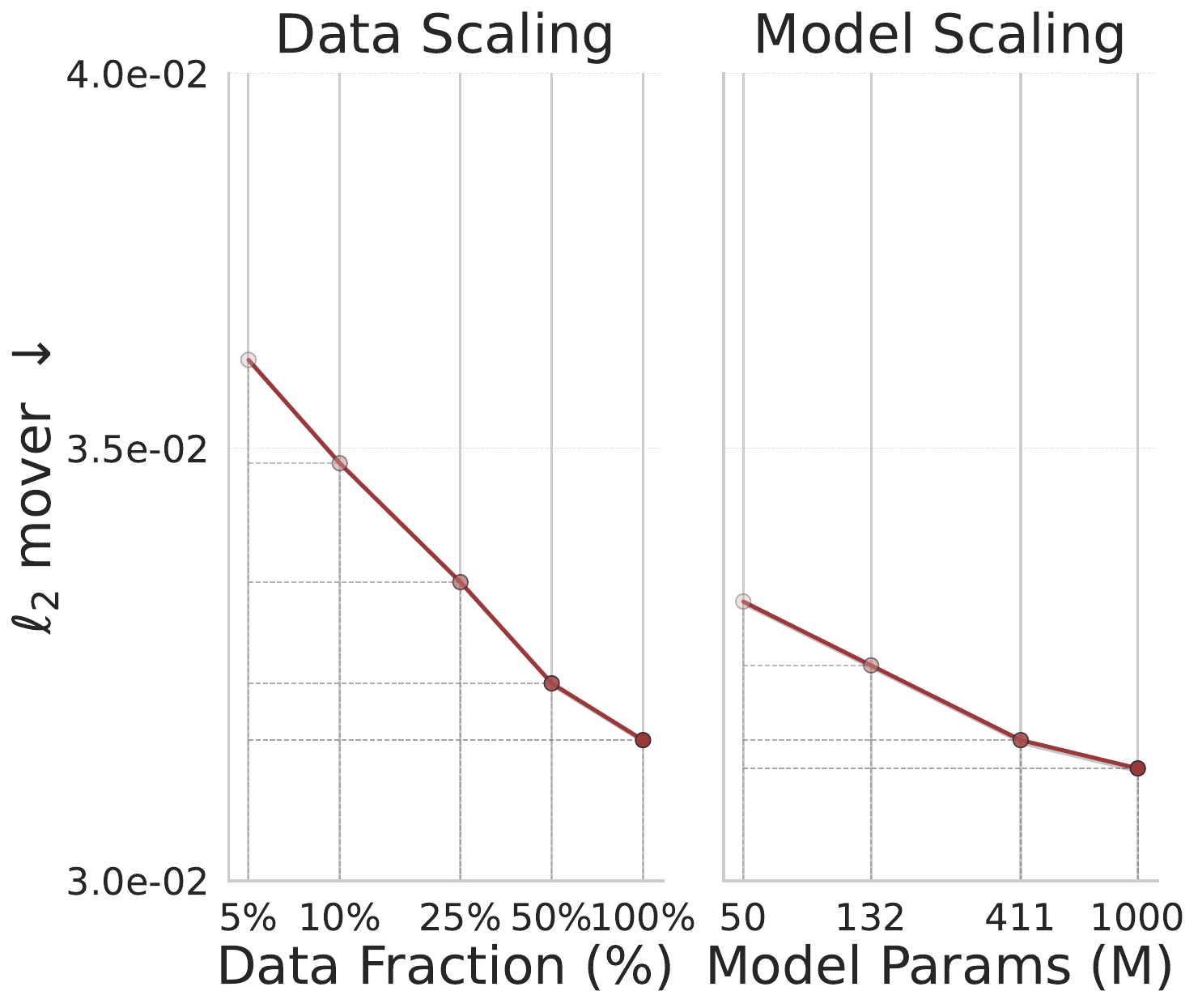}
        \captionof{figure}{\textbf{Scaling Study}. Scaling \algo in either data or model size yields roughly log-linear gains in prediction accuracy.}
        \label{fig:scaling-law}
    \end{minipage}
    \vspace{-0.8em}
\end{figure*}

\subsection{Scaling 3D World Models: A Roadmap}

\textbf{Modern point cloud backbone (PTv3~\cite{ptv3}) is effective, efficient, and scalable for 3D world modeling.}
Graph-based neural dynamics (GBND) models are widely used for dynamics modeling due to their relational inductive bias \cite{ai2025_survey}.
Scaling a GBND baseline to our dataset reveals two challenges (Table~\ref{tab:backbone}).
Memory consumption grows rapidly because maintaining high-dim features for all points in a scene is expensive.
Purely local message passing struggles under partial observability, since long-range effects must traverse noisy hops.
Motivated by these limitations, we study alternative point cloud architectures, moving from PointNet~\cite{qi2017pointnet}, PointNet++~\cite{qi2017pointnet++}, sparse convolutional nets~\cite{spconv2022} to transformers~\cite{vaswani2017attention}.
Among these, PointTransformerV3~\cite{ptv3} (PTv3) delivers the strongest modeling power.
Its point serialization mirrors GBND's local grouping, while U-net hierarchy enables attention over progressively coarser point sets for long-range modeling and substantial parameter growth.
Table~\ref{tab:backbone} shows that it scales to $957\times$ GBND while keeping modest memory and runtime increases.
These results motivate PTv3 as the default backbone.

\textbf{Movement weighting, uncertainty regularization, Huber loss stabilize 3D world model learning on real-world data.}
Discussed in Section~\ref{sec:method}, naïve $\ell_2$ loss is hard to optimize because only a fraction of points move ($1$--$5\%$).
Noisy real-world data exacerbates this.
We therefore adopt movement weighting, uncertainty regularization, and a Huber loss on 3D residuals.
Movement weighting alone over-emphasizes noisy signals, but the uncertainty head and robust loss temper the weights and reduce overfitting.
Together, these changes stabilize training and improve accuracy relative to an unweighted $\ell_2$ baseline.

\textbf{Pre-trained 2D features offer critical priors and substantial gains.}
High-quality pretrained 3D representations remain scarce despite compelling 3D geometry.
Methods such as Sonata \cite{sonata} make encouraging progress but often lag behind in fine-grained scenes.
Following \cite{foundationstereo,vggt}, we hypothesize that dense features from DINOv3~\cite{dinov3} provide objectness priors without explicit segmentation.
We therefore project points into calibrated cameras and attach features from multiple layers from a frozen DINOv3.
This simple addition substantially boosts accuracy over the baseline.

\textbf{Model size scaling is necessary to ingest large-scale world modeling data.}
With architecture, objective, and features in place, we expand depth and width within the same PTv3 blueprint.
Aligned with scaling-law observations in vision and language modeling \cite{kaplan2020scaling}, scaling model size from 50M to 1B parameters yields smooth, log-linear gains (Figure~\ref{fig:scaling-law}) similarly for 3D world modeling.

\textbf{Taken together}, all these levers—backbone, training objective, pre-trained feature, and model scaling—yield substantial gains over the original GBND baseline~\cite{ai2025_survey}.

\label{sec:exp:roadmap}
\definecolor{transferShade}{HTML}{EDEDED}
\definecolor{scratchText}{HTML}{4F4F4F}

\begin{figure*}[t]
    \centering
    \includegraphics[width=0.98\textwidth]{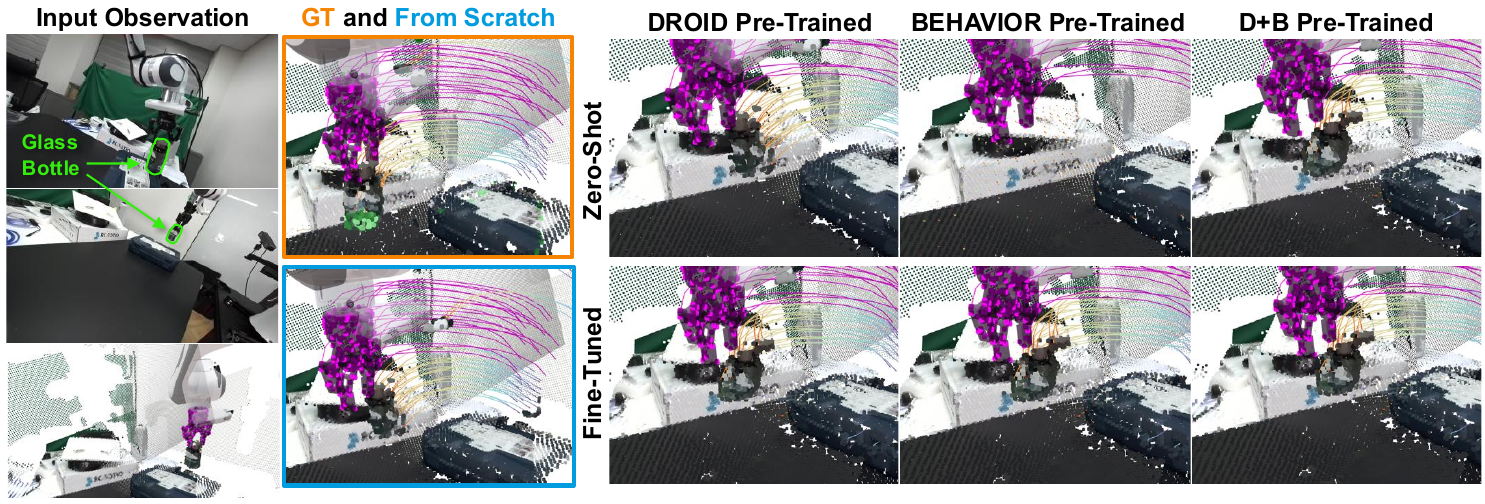}
    \vspace{-0.9em}
    \caption{\textbf{Zero-Shot and Finetuned Generalization to Held-Out Real-World Scenes}, where the robot transports a reflective glass bottle. \algo pre-trained on DROID or jointly on DROID and BEHAVIOR (D+B) are capable of zero-shot generalizing to unseen environment and motion from a held-out DROID lab's scene, closing the gap to the specialist variant trained on that lab's data. \algo pre-trained on only simulation data fail to generalize zero-shot. Further finetuning yields more accurate object trajectories of grasped objects.}
    \label{fig:zeroshot}
    \vspace{-0.7em}
\end{figure*}

\begin{figure*}[t]
    \centering
    \captionsetup{aboveskip=2pt, belowskip=2pt}
    \begin{minipage}[c]{0.62\textwidth}
        \centering
        {\footnotesize
        \setlength{\tabcolsep}{3pt}
        \renewcommand{\arraystretch}{1.2}
        \begin{adjustbox}{max width=\linewidth}
        \begin{tabular}{@{}l l c c|c c|c c c c@{}}
            \toprule
            & & \multicolumn{2}{c|}{In-Domain} & \multicolumn{2}{c|}{Cross-Domain} & \multicolumn{3}{c}{Held-Out Real} & \multicolumn{1}{c}{\textcolor{scratchText}{From}} \\
            & & $\text{D}\rightarrow\text{D}$ & $\text{B}\rightarrow\text{B}$ & $\text{D}\rightarrow\text{B}$ & $\text{B}\rightarrow\text{D}$ & $\text{D}\rightarrow\text{H}$ & $\text{B}\rightarrow\text{H}$ & $\text{D{+}B}\rightarrow\text{H}$ & \multicolumn{1}{c}{\textcolor{scratchText}{Scratch}} \\
            \midrule
            \multirow{2}{*}{\textbf{Zero-Shot}} & $\ell_2$ mover$\,\downarrow$ & 0.0315 & 0.0087 & 0.1460 & 0.0558 & 0.0305 & 0.0531 & 0.0300 & \textcolor{scratchText}{0.0293} \\
                & $\ell_2$ static$\,\downarrow$ & \cellcolor{transferShade}0.0059 & \cellcolor{transferShade}0.0010 & \cellcolor{transferShade}0.0050 & \cellcolor{transferShade}0.0058 & \cellcolor{transferShade}0.0049 & \cellcolor{transferShade}0.0057 & \cellcolor{transferShade}0.0063 & \cellcolor{transferShade}\textcolor{scratchText}{0.0043} \\
            \midrule
            \multirow{2}{*}{\textbf{Finetuned}} & $\ell_2$ mover$\,\downarrow$ & -- & -- & 0.0107 & 0.0378 & 0.0271 & 0.0299 & 0.0272 & \textcolor{scratchText}{0.0293} \\
                & $\ell_2$ static$\,\downarrow$ & \cellcolor{transferShade}-- & \cellcolor{transferShade}-- & \cellcolor{transferShade}0.0003 & \cellcolor{transferShade}0.0086 & \cellcolor{transferShade}0.0040 & \cellcolor{transferShade}0.0046 & \cellcolor{transferShade}0.0040 & \cellcolor{transferShade}\textcolor{scratchText}{0.0043} \\
            \bottomrule
        \end{tabular}
        \end{adjustbox}}
        \vspace{0.3em}
        \captionof{table}{\textbf{Generalization of \algo across in-domain, cross-domain, held-out real environments under zero-shot and finetuned settings.} $\text{D}$ denotes DROID, $\text{B}$ denotes B1K, $\text{H}$ denotes held-out real-world scenes.
        ``From Scratch'' denotes specialist trained on the held-out lab's data.
        Evaluations are done on unseen samples from the corresponding dataset.
        \algo generalizes within domains, zero-shot transfers to unseen real-world environments, surpasses specialists if finetuned with 20x fewer updates, and benefits from real-sim co-training.
        }
        \label{tab:domain-transfer}
    \end{minipage}
    \hfill
    \begin{minipage}[c]{0.36\textwidth}
        \centering
        \includegraphics[width=\linewidth]{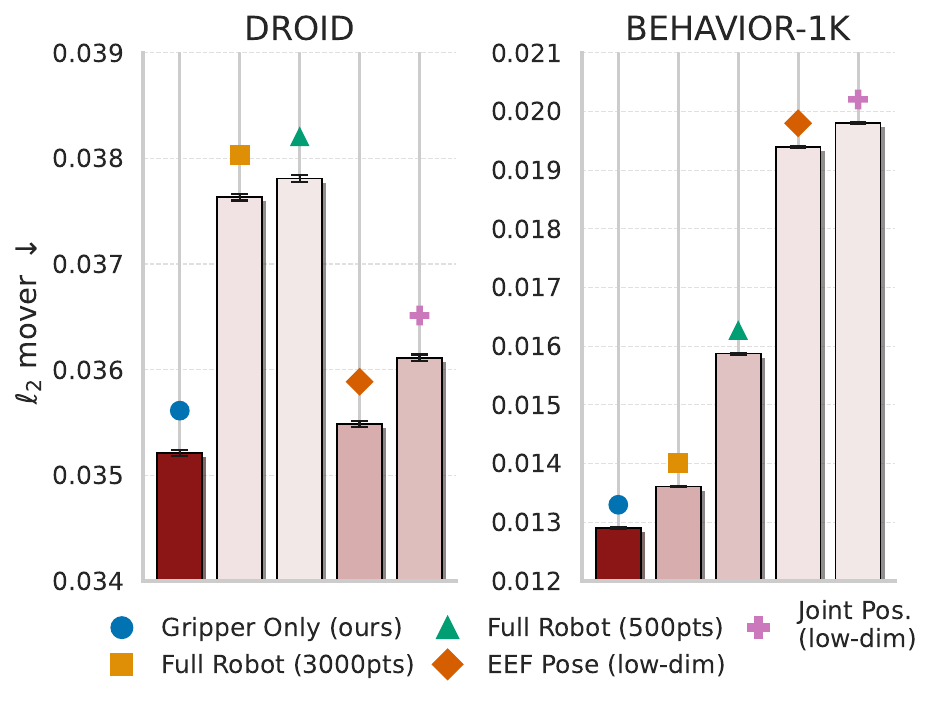}
        \vspace{-1.2em}
        \captionof{figure}{\textbf{Action representations.} Representing actions as point flows on grippers balances effective, efficient contact reasoning and enables positive transfer across heterogeneous embodiments.}
        \label{fig:action-representation}
    \end{minipage}
    \vspace{-1.9em}
\end{figure*}

\subsection{Ablations}
\label{sec:exp:ablations}

\textbf{Representing actions as point flows on grippers balances effective, efficient contact reasoning and enables positive transfer across heterogeneous embodiments.}
\label{sec:exp:action}
In \algo, robot actions are dense point flows over grippers with $300$--$500$ points per gripper.
We compare against four baselines:
(i) whole-body point flows with the same number of points (sparser coverage),
(ii) whole-body point clouds with $2000$ points (similar density as ours),
(iii) 6-DoF end-effector pose and gripper openness,
and (iv) joint positions and gripper openness.
The last two low-dim variants omit robot points, which the flow-based models concatenate with scene features.
We train all models jointly on both DROID and B1K data, where DROID uses a single-arm Franka and B1K uses a bimanual humanoid.
Results are in Figure~\ref{fig:action-representation}.
On B1K, representing contact spatially lets point-flow actions outperform low-dim alternatives (end-effector poses and joint positions).
Sparse whole-body flows underperform gripper-only flows, likely due to insufficient resolution to capture precise contact.
Dense whole-body flows help but still lag behind, as gradients must pass through inactive points and incur compute overhead.
On real-world DROID, both whole-body point-flow baselines underperform low-dim baselines.
A plausible explanation is that extensive robot points obscure already-sparse learning signals from noisy real-world data.
Gripper-only flows address this issue and attain the best performance, underscoring their effectiveness on real-world data and their ability to obtain positive transfer across heterogeneous embodiments in both domains.

\begin{figure}[t]
    \centering
    \includegraphics[width=\linewidth]{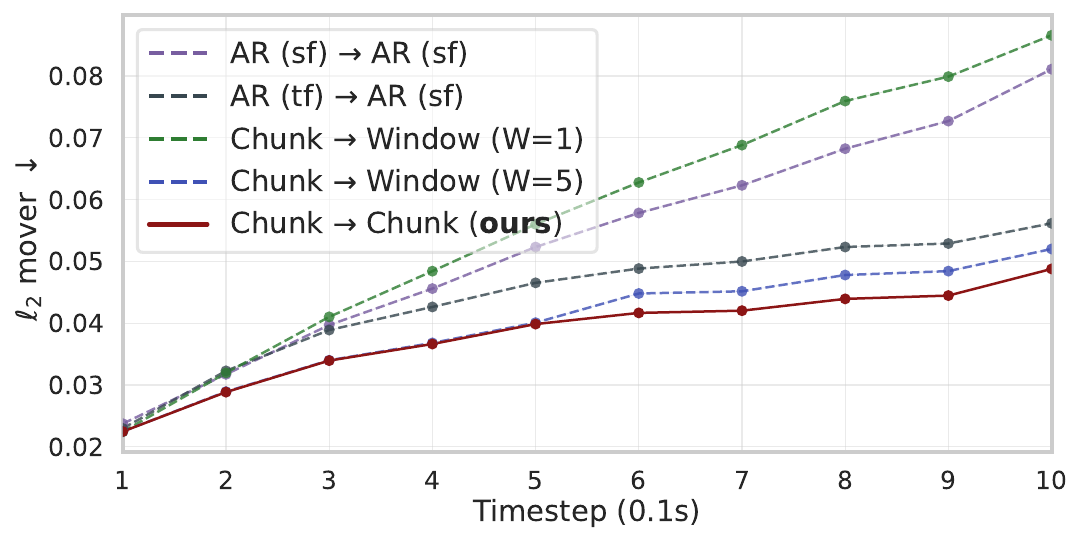}
    \vspace{-2.2em}
    \caption{\textbf{Ablation on Chunked Prediction}, where we study different rollout strategies in training and testing. 
    Chunked rollouts at both training and inference time lead to significantly less drift than other baselines while amortizing compute with only a single forward pass of the model.
    }
    \vspace{-0.7em}
    \label{fig:chunk-prediction-main}
\end{figure}

\textbf{Using chunked prediction in both training and inference reduces rollout drift while improving compute efficiency.}
\algo performs 10-step chunked prediction (equivalent to 1 second). We ablate this design choice against two autoregressive baselines: (i) teacher-forcing (GT input each step) and (ii) self-feeding with $10$k warmup steps, plus sliding-window inference ($W{=}1,5$) using the same chunked model.
Results are shown in Figure~\ref{fig:chunk-prediction-main}.
Teacher-forcing outperforms self-feeding when training and inference strategies are aligned.
Evaluating a chunk-trained model with $W{=}1$ (equivalent to self-feeding) incurs the strongest performance degradation; $W{=}5$ recovers some accuracy but degrades after the trained window.
Matching chunked prediction in training and testing over the full horizon minimizes drift while amortizing compute with only a single forward pass (vs.\ 2--10 for autoregressive), highlighting chunking as both more accurate and more compute-efficient design choice.

\begin{figure}[t]
    \centering
    \includegraphics[width=\linewidth]{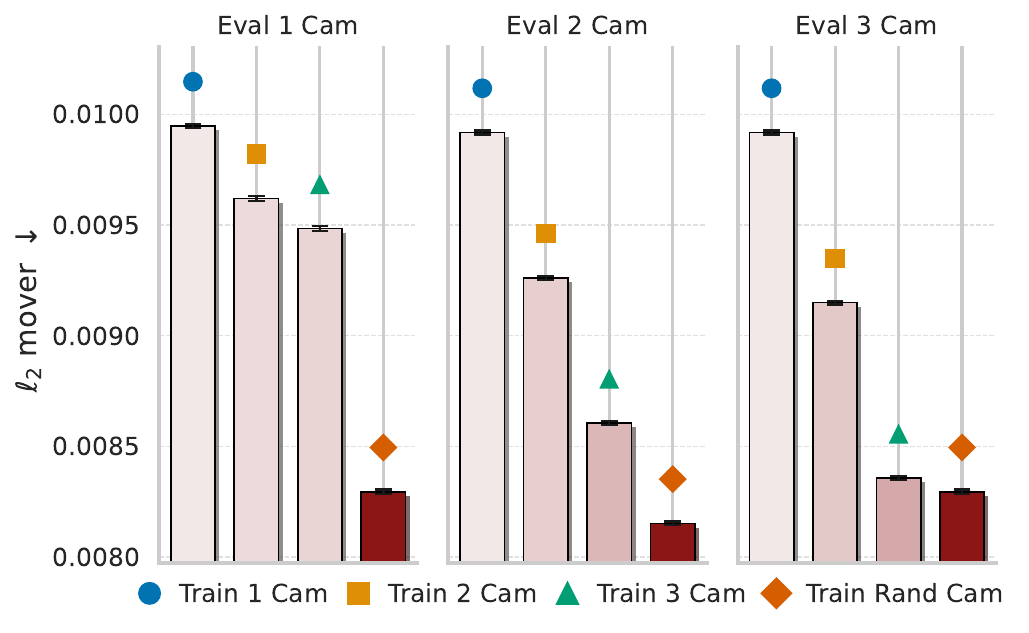}
    \vspace{-2.2em}
    \caption{\textbf{Ablation on Partial Observability}, where we train variants of \algo with varying number of cameras and evaluate them on all settings at test time.
    \algo is robust to different levels of partial observability and benefits from additional cameras in both training and inference.
    Training with randomized camera counts yields the best performance across all test settings.
    }
\label{fig:number-of-views-main}
    \vspace{-1.1em}
\end{figure}

\textbf{\algo is robust to different levels of partial observability and benefits from additional cameras in both training and inference.}
We train four variants that differ only in camera count for RGB-D observations: one, two, three, or a random draw of up to three cameras.
We then evaluate all models on three settings with up to three cameras.
Results are in Figure~\ref{fig:number-of-views-main}.
Error on moving points stays sub-centimeter with negligible standard errors, but using more cameras at train time consistently reduces error at test time.
Interestingly, models trained with fixed camera count perform better when more cameras are available at inference.
The random-view model is most robust across all test camera counts, suggesting that exposure to varied observability helps the model infer objectness and physical properties under partial observability at inference time.

\textbf{Prediction error decreases roughly log-linearly with both model size and data.} Inspired by scaling laws from language and vision \cite{kaplan2020scaling,chowdhery2022palm,alayrac2022flamingo}, we test whether \algo follows similar trends. On DROID, we vary model capacity (50M--1B) and data fraction (5\%--100\%). Each curve sweeps one axis only. In log space we observe approximately linear behavior for both axes (Figure~\ref{fig:scaling-law}), suggesting predictable gains from extra data and capacity.

\subsection{Generalization and Transfer}
\label{sec:exp:generalization}
		
We study \algo's generalization across in-domain, cross-domain, and to held-out real-world environments under zero-shot and finetuned settings.
Each finetuning uses $1/20$ of the original training iterations.
Results are in Table~\ref{tab:domain-transfer}.
	
\textbf{\algo generalizes within domains.}
We study in-domain transfer on held-out splits of DROID and B1K that are unseen during training.
On B1K the model achieves sub-centimeter mover error on held-out trajectories, while DROID performance on held-out remains similar to training despite real-world variations. This indicates that \algo does not simply memorize training samples.
	
\textbf{Pre-trained \algo can be efficiently finetuned (20x fewer updates) for both real-to-sim and sim-to-real transfer.}
We study cross-domain transfer by evaluating how a model pre-trained on DROID generalizes to B1K, and vice versa.
Zero-shot transfer between simulation and real domains remains challenging.
Yet, finetuning with only $5\%$ of the original training steps rapidly narrows the gap to domain-specific models trained from scratch using $20\times$ more updates. The effect is symmetric: real-to-sim and sim-to-real transfers both benefit.
Empirically, we observe training on real-world data provide better transfer than reverse, plausibly due to the higher scene diversity of the real-world data.
	
\textbf{\algo zero-shot generalizes to unseen real-world environments, surpasses specialists if finetuned with 20x fewer updates, and benefits from real-sim co-training.}
To study held-out real generalization, we hold out data from the CLVR lab within DROID and evaluate how well a model pre-trained on the remaining DROID data generalizes to that lab.
The held-out set is split into $90\%$ train and $10\%$ test. Zero-shot models never see these frames, while finetuned variants access only the $90\%$ subset.
\algo pre-trained on the remaining DROID data achieves on-par performance with specialists trained on the held-out lab despite changes in background, lighting, object, and possibly motion distribution.
With finetuning, it quickly surpasses the specialist.
We observe simulation-pretrained models do not outperform scratch baselines yet reach comparable accuracy with finetuning.
Finally, a model pre-trained on combined DROID and B1K mix delivers mildly stronger zero-shot performance than DROID-only.

\FloatBarrier

\subsection{Model-Based Planning with \textbf{\algo}}
\label{sec:exp:robotics}
Pre-trained on diverse interactions, we test whether \algo can be zero-shot deployed for manipulation on a physical robot in the wild.
For evaluation, we use a Franka setup similar to DROID, mounted on a wheeled base and equipped with one RealSense D435 camera.
Depth is estimated using FoundationStereo~\citep{foundationstereo}.
For each task, we manually draw an object mask and specify target positions through a GUI tool.
Each optimization rolls out 30 steps (3 autoregressive forward passes).
With only the pre-trained model and a shared MPC framework, \algo optimizes actions for real‑world tasks: non‑prehensile pushing of rigid objects (tissue box, book), deformable manipulation (folding a scarf, placing a pillow), articulated manipulation (opening a microwave and closing a drawer, with revolute and prismatic joints), and tool use (sweeping with a duster or broom).
Tasks and success rates are shown in Figure~\ref{fig:robot-success},
indicating the pre-trained \algo captures transferable interaction dynamics,
including contact reasoning under partial observability (rigid pushing),
implicitly inferring articulation and deformation of objects (articulated and deformable manipulation),
and object-object interactions (tool use).

\FloatBarrier
\section{Conclusion}
We introduced \algo, a large pre-trained 3D world model, that predicts 3D environment dynamics given in-the-wild RGB-D capture(s) and robot actions under a shared representation of 3D point flows.
To train the model, we leveraged recent advances in 3D vision and curated a large-scale dataset for action-conditioned 3D world modeling, with high-quality depth maps, camera poses, and 3D tracks.
Through empirical evaluations, we rigorously studied the recipe for scaling 3D world model training, including backbone designs, action representations, learning objectives, partial observability, data mixtures, domain transfers, and scaling laws.
Pre-trained on diverse data, a single \algo model enabled practical manipulation behaviors in the real world, including non-prehensile pushing, deformable and articulated object manipulation, and tool use.

\section*{Acknowledgments}
This work is in part supported by the Stanford Institute for Human-Centered AI (HAI), the Schmidt Futures Senior Fellows grant, ONR MURI N00014-21-1-2801, ONR MURI N00014-22-1-2740.
We would like to thank Leslie Kaelbling, Tomás Lozano-Pérez, Yunzhu Li, Jiajun Wu, Ruohan Zhang, Jiayuan Mao, Abhishek Gupta, Pieter Abbeel, Pulkit Agrawal, Shenlong Wang, Anirudha Majumdar, Chuang Gan, David Held, Karen Liu, Jeannette Bohg, Balakumar Sundaralingam, Anqi Li, Bowen Wen, Xiaoyang Wu, Hang Gao, Chelsea Ye, Mijiu Mili, the BEHAVIOR team, members of the Stanford Vision and Learning Lab, and members of the Learning \& Intelligent Systems Group for fruitful discussions, feedback, and support.
\FloatBarrier
{\small
\bibliographystyle{ieeenat_fullname}
\bibliography{main}
}

\appendix
\clearpage
\section{Appendix}
\label{sec:appendix}

\begingroup
    \setcounter{tocdepth}{3}%
    \etocsettocstyle{\section*{Appendix Contents}}{}
    \localtableofcontents
\endgroup
\raggedbottom

\makeatletter
\newcommand{\appclearpage}{%
    \clearpage
    \if@twocolumn
        \global\@firstcolumntrue
    \fi
    \global\@textfloatsheight\z@
    \global\@colroom\@colht
    \global\vsize\@colroom
}
\makeatother

\appclearpage
\subsection{Extended Discussions on Limitations}

\paragraph{Static Initial State.}
The model takes as input RGB-D point cloud together with a finite-horizon sequence of robot actions and predicts how points will move in response.
Because no prior frames or velocities are provided, this formulation assumes that the world is static at the observation instant.
Supporting fully dynamic initial conditions would require augmenting the input with externally tracked trajectories or recurrent state, which we leave as future work.

\paragraph{Reward/Cost Specification for Action Inference.}
In this work, we explore \algo's use case for action inference in manipulation by integrating it with a sampling-based planner, MPPI~\cite{mppi}, which requires an explicit specification of reward/cost functions in the same state-action space of 3D point flows.
For the scope of this work, we restrict ourselves to manual specification (e.g., moving a subset of points to target locations).
Future work may consider automatically specifying (single or multi-stage) reward via vision-language models~\cite{huang2024rekep}, or inferring reward from demonstrations using inverse reinforcement learning~\cite{abbeel2004apprenticeship}, while keeping \algo as the dynamics function.
In addition to planning, action inference can also be alternatively performed by learning a parameterized policy by interacting with the model as the environment via reinforcement learning~\cite{dreamer}.

\paragraph{Fine-Scale Objects and Calibration Noise.}
Thin or very small objects (e.g., pens or cables) are challenging to annotate accurately in 3D: modest depth or extrinsic errors can be comparable to the object thickness and lead to ambiguous separation between robot and scene points during ground-truth annotation.
In such scenarios, mis-registrations in the ground-truth flows propagate into training and can cause the model to confuse overlapping motions between the gripper and nearby scene points.
Improved calibration and depth estimation could strengthen supervision for these fine-grained interactions.

\paragraph{Correlation vs.\ Causation.}
Given an observed context frame and a sequence of robot actions, \algo is trained to predict the subsequent sequence of scene states.
As such, it primarily captures correlations present in the training distribution between robot actions, robot motion, and observed scene evolution.
In settings where exogenous factors (such as other agents or environment changes not controlled by the robot) also influence the future, these influences are entangled with the robot-induced effects in the data and are not disentangled as separate causal mechanisms.
Our experiments therefore evaluate predictive fidelity and planning performance under the observed action-conditioned distribution, rather than claiming to recover the underlying causal structure of the environment.

\paragraph{Lack of Photometric Dynamics.}
\algo only outputs displacements of 3D points captured from RGB-D inputs, which focus on geometry and physical interactions rather than appearance.
While often visually plausible when rendered as point clouds, it is insufficient if one desires to reason about photometric changes of the environment caused by the robot actions, such as lights or screens turning on and off.
Combining \algo with appearance models that predict emitted radiance, such as those from Gaussian Splatting~\citep{gaussian_splatting} or Neural Radiance Fields~\citep{nerf}, may be necessary for tasks where such photometric dynamics are critical.

\paragraph{Rigid-Body Robot Assumption.} Robot embodiment is represented as a kinematic tree of rigid links, and we propagate a fixed set of robot surface points by forward kinematics. This ignores deformations of soft, tendon-driven, or compliant structures (e.g., fin-ray grippers) and non-rigid effects of the robot body. As a result, \algo reasons only about how the scene moves in response to a forecasted robot geometry, rather than how contact may reshape the robot itself. Extending the representation to include deformable links~\cite{li2025controlling} would enable reasoning about how contact deforms the robot body and how those deformations, in turn, affect contact geometry.

\paragraph{Actuation and Tracking Assumptions.} Our formulation treats the robot trajectory as a known, \emph{fully realized} sequence of joint configurations and uses forward kinematics to construct robot flows. As a result, \algo effectively models “what the environment does if the robot body follows this path,” rather than “whether and how the robot will actually realize this path” under a particular controller, actuation limits, or contact-induced tracking errors. This quasi-static, kinematic view of the robot action representation can break down for underactuated or compliant joints (e.g., tendon-driven or compliant hand fingers), or when strong contacts, payloads, or controller changes induce large tracking errors. Extending the method to jointly reason about robot and scene dynamics is an important avenue for future work.

\paragraph{Lack of Explicit Physics Priors.}
The current formulation is purely data-driven and does not incorporate explicit physics priors such as Newtonian mechanics or conservation-law constraints, to provide a focused study on scaling 3D world models without priors of objectness and of their material and physical properties.
Despite this, we observe that \algo captures many aspects of rigid, articulated, and deformable behavior from data alone.
Incorporating physics-informed regularization or hybrid simulators~\citep{phystwin2025} could further improve generalization and extrapolation, particularly in regimes that in-domain interaction data can be collected for accurate scene/object reconstruction, not only for their geometries but also for their physics parameters.

\appclearpage
\subsection{DROID 3D Annotation Pipeline}
\label{app:real-data}

DROID~\citep{droid} is a large-scale robot manipulation dataset with human-teleoperated interactions collected with a wrist-mounted camera and two externally-mounted cameras (randomized over the left and right sides of the workspace).
We use all DROID episodes for which raw camera streams are available, irrespective of task success or failure, since 3D world modeling depends only on the observed interactions rather than the task-specific outcomes of the manipulation.
Each episode provides stereo RGB streams with ground-truth camera intrinsics for all three cameras, plus robot joint states and a known kinematic model of the robot.
In this work, we use the recovered 3D scene flows from the two externally-mounted cameras.
All data share a synchronized timestamp.
We augment the robot model to include the Robotiq 2F-85 gripper and the custom camera mount used in the standardized DROID setup.
For each scene, the pipeline aligns timestamps to a reference stream (binary search to nearest), downsamples by 2, then runs: (i) dense metric depth; (ii) external-camera extrinsic refinement by aligning rendered robot mesh to observed depth; (iii) 2D tracking under workspace and robot masks; (iv) 3D trajectory reconstruction, slicing, and postprocessing.
Note that we do not store robot point flows because those can be reconstructed efficiently at training/inference time given known robot URDF and the given joint actions.

\subsubsection{Depth Estimation}
Per-view metric depth is obtained using a high-quality stereo estimator, FoundationStereo~\citep{foundationstereo}.
Note that unlike typical sensor depth, the estimated depth from FoundationStereo does not have a minimum depth threshold for valid depth perception.
However, it is still observed that its estimated depth can be inaccurate for distant, especially texture-less, regions (such as walls).
Therefore, depth values are sanitized by clamping to a trusted range \([0, 4]\,\mathrm{m}\) and producing a per-pixel validity mask, which is also propagated to 3D points as a per-point depth-valid flag.

\subsubsection{Camera Pose Estimation}
\label{app:data:extrinsics}
We do not use dataset-provided extrinsics.
Instead, we compute camera extrinsics using a two-stage procedure that leverages the accurate metric depth obtained from FoundationStereo discussed previously.
First, we initialize the camera pose estimates using VGGT~\citep{vggt}.
Second, we refine all camera poses of the two external cameras from all timesteps jointly by aligning rendered robot geometry to observed depth using the recorded robot joint states.

\paragraph{Camera Pose Initialization.}
Our goal is to estimate, for each externally mounted camera $C_i$, a single rigid transform $T_{C_i \leftarrow B} \in \mathrm{SE}(3)$ that maps 3D points from the robot base frame $B$ into the camera frame and is fixed throughout a DROID episode.
We denote the robot base frame by $B$, the wrist camera frame at time $t$ by $W_t$, and the external cameras by $C_i$.
A multi-view pose estimator (VGGT~\citep{vggt}) is applied to time-aligned images from the two external cameras and the wrist camera; it treats the first external camera at an initial timestep as the reference frame, and returns rigid transforms $T_{E_0 \leftarrow C_i}$ and $T_{E_0 \leftarrow W_t}$ that map points from each camera $C_i$ or $W_t$ into this reference frame $E_0$.
Independently, forward kinematics applied to the robot joint states provide the pose of the gripper in the base frame, $T_{G_t \leftarrow B}$.
For each physical robot (given by the recorded robot serial), we assume that the wrist camera is rigidly mounted relative to the end effector and reused across all episodes.
We empirically found that this assumption is largely valid for robots used in DROID as the averaged transforms exhibit sub-millimeter alignment error with each other.
Under this assumption, we can obtain a known gripper-to-wrist transform $T_{W \leftarrow G}$ for each robot.
Using this transform and the forward-kinematics model, we obtain a time-varying wrist pose in the base frame,
\[
    T_{W_t \leftarrow B} = T_{W \leftarrow G}\, T_{G_t \leftarrow B}.
\]
Combining this with the estimator’s transform between the wrist and the reference external camera yields per-frame estimates of the reference camera pose in the base frame,
\[
    T_{E_0 \leftarrow B}^{(t)} = T_{E_0 \leftarrow W_t}\, T_{W_t \leftarrow B},
\]
which we average over all valid wrist frames to obtain a single $T_{E_0 \leftarrow B}$.
For any other external camera $C_i$, the estimator provides a relative transform $T_{E_0 \leftarrow C_i}$.
We convert this to a base-frame extrinsic
\[
    T_{C_i \leftarrow B} = T_{C_i \leftarrow E_0}\, T_{E_0 \leftarrow B}, \quad T_{C_i \leftarrow E_0} = T_{E_0 \leftarrow C_i}^{-1},
\]
so that all external cameras are expressed in the same robot base frame before the refinement stage.

\paragraph{Camera Pose Refinement.}
Starting from the initialized base-frame extrinsics $T_{C_i \leftarrow B}$, we jointly refine the poses of all external cameras using robot-depth reprojection.
Let external cameras be indexed by $i$, timesteps by $t$, and let $k$ index valid robot pixels after filtering (in front of the camera, within image bounds, deduplicated in the image plane, and with observed depth in the trusted range).
For each camera $i$ we optimize a small 6-DoF update on top of the initialization, parameterized by translation and rotation and scaled such that optimization can be done within a good numerical range.
Given an observed depth value $d^{\text{obs}}_{i,t,k}$ at a valid robot pixel and the corresponding predicted depth $d^{\text{pred}}_{i,t,k}$ obtained by projecting robot surface points from the base frame through the current extrinsics $T_{C_i \leftarrow B}$, we define the robot-depth reprojection loss
\[
    L_{\text{robot-depth}} = \frac{1}{K} \sum_{i,t,k} \bigl| d^{\text{obs}}_{i,t,k} - d^{\text{pred}}_{i,t,k} \bigr|,
\]
where the sum runs over all valid robot pixels across cameras and frames and $K$ is their total count.
We optimize the 6-DoF updates for all external cameras jointly using a first-order optimizer (100 iterations, learning rate $10^{-3}$), and restrict supervision to robot pixels whose observed depth lies in a trusted range of $[0.3, 2.0]$\,m.
To ensure reliable gradients, we further require that each camera--frame pair contributes at least $2{,}000$ valid robot points; frames that fail this criterion for any external camera are discarded before refinement.
With these procedures, we can accurately label camera poses for around 60\% of all episodes recorded in DROID. Quantitative metrics are reported in the main text.

\subsubsection{Benchmark Metrics for 3D Annotation}
\label{app:data:metrics}
\paragraph{Depth Reprojection.}
For each scene and calibration variant, we render the robot mesh into each external camera using the candidate extrinsics and recorded joint states, discard depth outside $[0.3, 2.0]$\,m or out-of-bounds robot pixels, and compute an L1 depth reprojection loss over valid robot pixels.
The per-frame losses are averaged over valid pixels and frames to produce a point-weighted value reported per scene and then aggregated into cumulative curves.

\paragraph{Two-view F1 @ 5/20 mm.}
For each external camera, we back-project valid depth into 3D using corresponding intrinsics/extrinsics, remove robot pixels, and crop to workspace.
Given the resulting paired point clouds, we compute precision, recall, and F1 via symmetric nearest-neighbor matching: a point in cloud A (resp.\ B) is a true positive if its nearest neighbor in B (resp.\ A) lies within the threshold (5\,mm or 20\,mm); otherwise it contributes to the false-positive/false-negative count.
Metrics are accumulated over frames, normalized by the number of valid points per view, and then aggregated per scene into the cumulative counts shown in \figurename~\ref{fig:annotation}.
Scenes with missing depth/extrinsics for a given combo are omitted from evaluation for that combo.

\subsubsection{Occlusion-Aware Tracking in 3D}
Given depth and camera poses, we herein describe how we can obtain 3D point flows.
We describe the following in order: (i) filtering clips based on robot motion, (ii) tracking visible points within each retained clip, and (iii) postprocessing the resulting 3D trajectories.

\paragraph{Clip Filtering.}
We slice each episode into overlapping clips of length $F{=}16$ frames with stride $s{=}1$.
Each clip covers approximately one second of the episode.
Clips are retained if either the gripper changes state within the clip or end-effector motion exceeds thresholds.
Thresholds depend on whether the gripper is predominantly open or closed within the clip: position/rotation thresholds are $(0.005\,\mathrm{m},\,0.10\,\mathrm{rad})$ when open and $(0.002\,\mathrm{m},\,0.05\,\mathrm{rad})$ when closed; either exceeding suffices to keep a clip, and any change in gripper state also keeps the clip.

\paragraph{Tracking.}
After obtaining depth and camera poses and slicing episodes into short clips, we perform dense tracking to obtain 3D scene flows on a per-clip basis.
Clips cover roughly one-second windows and, depending on where they fall within a longer episode, may observe quite different regions of the workspace; tracking after clip selection ensures that each clip has its own consistent set of tracked regions and avoids mixing trajectories across widely separated time intervals.
To improve efficiency and robustness, we perform tracking restricted to workspace and non-robot regions.
To construct the workspace mask, we project a fixed 3D workspace volume to the image.
To construct the non-robot mask, we render the robot’s URDF and project the mesh to the image plane.
We then track 2D points using 2D point trackers (CoTracker3~\citep{karaev2025cotracker3}) on the masked regions only, producing dense 2D trajectories.
The tracker also outputs a per-point visibility mask over time; we store this visibility for each 2D trajectory so that occluded points can later be excluded from supervision after lifting to 3D.
Because \algo is trained on 3D point flows rather than image-plane trajectories, we lift each tracked 2D point to a 3D world-space trajectory by back-projecting it with the corresponding depth, intrinsics, and extrinsics at each timestep.
We next reconstruct 3D trajectories and apply clip-level postprocessing.
For each frame, valid depth, intrinsics, and extrinsics back-project tracked pixels to world-frame 3D points with RGB.
Tracks across time yield temporally consistent per-point trajectories.
We store per-camera trajectories to avoid mixing viewpoints prematurely and keep per-point visibility and depth-valid flags that are later used to mask supervision.

\paragraph{Postprocessing.}
To further improve the quality of the obtained 3D point flows, we apply two postprocessing steps: DBSCAN-based outlier removal and per-frame normal estimation.
For each clip we first remove spatial outliers using multi-scale DBSCAN clustering~\citep{ester1996density} across all external cameras: at each timestep, we run DBSCAN with radii $\varepsilon\in\{0.02, 0.05\}$\,m and minimum core size $5$, and mark point flows that are classified as outliers in more than $20\%$ of frames as outliers to be discarded.
From the remaining trajectories, we estimate normals per frame using local neighborhoods (up to $k{=}30$ neighbors within radius $0.1$\,m) and orient them toward the camera, followed by a temporal consistency step that flips back-facing normals so that normal directions remain coherent over time.

\appclearpage
\subsection{BEHAVIOR-1K Data Generation}
\label{app:data-sim}
BEHAVIOR-1K (B1K)~\citep{behavior} is a large-scale benchmark of everyday household activities in photorealistic simulation built on NVIDIA Isaac Sim.
As part of the 2025 BEHAVIOR Challenge, it provides approximately $10{,}000$ human-teleoperated episodes (average length $\approx 6.6$ minutes) across $50$ tasks executed by a bimanual mobile robot (Galexea R1 Pro).
We replay these episodes in the original simulator and attach three virtual cameras (left shoulder, right shoulder, and head) to extract short clips with meaningful interactions and dense 3D point flows, as detailed below.

\subsubsection{Dataset Replay}
\label{app:data-sim:replay}
For each BEHAVIOR-1K episode, we replay it in the simulator using the recorded sequence of environment states and actions.
To prevent physics leakage and adhere to the original demonstrations, we iterate over the stored trajectory and, at every recorded step, load the corresponding simulator state and advance the simulator once with the recorded action.
At every such step, we also render three external RGB-D cameras mounted on the robot: a left and right shoulder camera attached near the base, and a head-mounted camera.
All three cameras have ground-truth intrinsics and extrinsics, and produce per-pixel depth, surface normals, and per-link segmentation in addition to RGB.
All extrinsics are recorded in the robot base frame in the first timestep of each clip.

\subsubsection{Clip Filtering}
\label{app:data-sim:clips}
We aim to extract short clips of fixed length $F{=}11$ frames that contain meaningful interaction between the robot and the scene while discarding static or uninteresting intervals.
To generate candidates, we slide an overlapping window of length $F$ over each replayed episode at a fixed temporal stride; any window for which \emph{all} external cameras have no visible, in-workspace scene objects---that is, no non-robot, non-ground meshes with valid depth inside the workspace bounds—is immediately discarded.

For each remaining candidate window, we maintain a set of \emph{motion indicators} and \emph{contact indicators} that are updated over the clip.
Let $M_g$ denote the event that at least one arm's end-effector exhibits sufficient translational or rotational motion over the clip or undergoes a change in gripper open/closed state, with thresholds that depend on whether the gripper is predominantly open or closed.
Let $M_j$ denote the event that any non-base robot joint moves more than a fixed threshold over the clip.
Using the object trajectories described in Section~\ref{app:data-sim:flows}, we define $M_o$ as the event that at least one object moves more than an object-movement threshold in position or orientation relative to its pose at the first frame of the clip.

From the ground-truth simulation state, we further construct contact indicators.
Let $C_t$ denote the event that any trunk or arm link experiences a nonzero contact impulse during the clip, and let $C_f$ denote the event that any gripper finger link experiences contact.
Clips that contain large simulator-induced discontinuities (such as scene resets) are filtered internally before applying the following logical criterion.

At the end of the horizon, a remaining candidate is accepted as a valid clip if and only if
\begin{equation}
\label{eq:clip-validity}
    \neg C_t \land \bigl((M_o \land M_j) \lor (M_o \land C_f) \lor (\neg M_o \land M_g \land M_j)\bigr).
\end{equation}
The term $\neg C_t$ discards clips that contain trunk or arm collisions.
The first disjunct in~\eqref{eq:clip-validity} retains clips where object motion is causally associated with non-base joint motion.
The second disjunct retains clips where object motion primarily arises through gripper-finger contacts, which covers behaviors such as pushing an object with only base motion rather than arm motion (e.g., pushing a door by moving the base).
The third disjunct retains ``negative'' clips in which the robot moves but no objects move, providing supervision on background dynamics and free-space motions.

\subsubsection{3D Point Flows from Simulation}
\label{app:data-sim:flows}
For each accepted clip and each external camera, we construct a compact representation of 3D point flows that exploits three properties of the simulation setting:
(i) the environment is composed of rigid objects decomposed into rigid links;
(ii) we have access to ground-truth link-level instance segmentation in rendered images; and
(iii) we can query the exact rigid trajectory of every link throughout the clip.
At the first frame of the clip, we back-project depth for each visible link to obtain a set of surface points in that link's local frame, together with associated colors and normals, after filtering out background and robot meshes and enforcing workspace bounds in the robot base frame at the clip start.
We then record the time-varying poses of all visible links and cameras in this same clip-start robot base frame.
This factorized representation, local link points plus per-link trajectories, allows us to reconstruct exact per-point 3D trajectories for any clip while remaining far more storage-efficient than storing dense point clouds at every frame.
Note that while we access ground-truth simulator state for obtaining ground-truth 3D point flows, the simulator state is never exposed to the model.

\begin{table*}[t]
    \centering
    \setlength{\tabcolsep}{4pt}
    \small
    \begin{tabularx}{0.9\linewidth}{l X}
        \toprule
        \textbf{Operation} & \textbf{Description} \\
        \midrule
        Camera subsampling & Sample randomized calibrated RGB-D views per timestep and concatenate their 3D points into a single scene cloud. \\
        Bounds filtering & Retain only scene points that stay within a workspace cube (approx.\ \([-3,3]^3\) m) for the entire clip, dropping particles that ever exit the bounds. \\
        Centering & Mean-center first-frame scene and robot points. \\
        Image resize & Downscale RGB-D images to $320\times180$. \\
        Voxel downsampling & Voxel-grid sampling at $1.5$\,cm; select one point per occupied voxel at $t{=}0$, and apply the same indices to all timesteps. \\
        Multi-sphere cropping & Iteratively remove spheres of points far from the robot (up to three spheres, radii in $[0.10,0.80]$\,m with buffer $0.25$\,m) until the scene falls below the target budget. \\
        Max scene / robot points & Randomly subsample scene points if more than $12\,000$ remain after cropping; robot points are capped at $500$ by construction. \\
        Random yaw & Uniform rotation about the vertical axis over $[-\pi,\pi]$. \\
        Uniform scaling & Isotropic scaling with factor sampled uniformly from $[0.9,1.1]$. \\
        Random reflection & With probability $0.5$, reflect the scene and robot across either the $x$- or $y$-axis. \\
        Chromatic auto-contrast & Apply chromatic auto-contrast to RGB channels with probability $0.2$ and blend factor up to $0.2$. \\
        Chromatic translation & Add a global RGB offset with magnitude 2\% with probability $0.95$. \\
        Chromatic jitter & Add per-point RGB noise with standard deviation 2\% with probability $0.95$. \\
        \bottomrule
    \end{tabularx}
    \caption{\textbf{Data Preprocessing and Augmentations.}}
    \label{tab:augs}
\end{table*}

\begin{table*}[t]
    \centering
        \setlength{\tabcolsep}{4pt}
        \small
        \begin{tabularx}{0.9\linewidth}{l l X}
            \toprule
            \textbf{Point set} & \textbf{Feature} & \textbf{Definition} \\
            \midrule
            Robot & Position $p^{\text{robot}}_{t,j}$ & 3D coordinates of robot points over time. \\
            Robot & Color $c^{\text{robot}}_{j}$ & Constant magenta color \((1,0,1)\) indicating robot identity, shared across timesteps. \\
            Robot & Normal $n^{\text{robot}}_{t,j}$ & Surface normals of robot points from the known robot URDF. \\
            Robot & Gripper openness $\tilde{g}_t$ & Scalar gripper open value per timestep, broadcast to all robot points. \\
            Robot & Velocity $v^{\text{robot}}_{t,j}$ & Per-point velocity from mid-point finite differences over $p^{\text{robot}}_{t,j}$ across time. \\
            Robot & Acceleration $a^{\text{robot}}_{t,j}$ & Per-point acceleration from mid-point finite differences over $v^{\text{robot}}_{t,j}$ across time. \\
            Scene & Position $x_{0,i}$ & 3D coordinates of scene points at the first frame after preprocessing. \\
            Scene & Color $c^{\text{scene}}_{0,i}$ & RGB color of scene points at the first frame. \\
            Scene & Normal $n^{\text{scene}}_{0,i}$ & (Estimated) surface normals of scene points at the first frame. \\
            Scene & Gripper openness sequence $g_{0{:}T-1}$ & Sequence of gripper openness values over the context and prediction horizon, broadcast to every scene point. \\
            Scene & Distance-to-robot $d_{0{:}T-1,i}$ & For each timestep, distance from the first-frame position of scene point $i$ to the closest robot point, stacked across time. \\
            \bottomrule
        \end{tabularx}
\caption{\textbf{Per-Point Input Features.}}
\label{tab:features}
\end{table*}

\appclearpage
\subsection{Model Training Details}
\label{app:model-config}

\paragraph{Data Preprocessing and Augmentations.}
Here we describe the data preprocessing and augmentations used in our experiments.
Each training sample fuses calibrated RGB-D views before passing through workspace filtering, centering, and deterministic voxel sampling with multi-sphere cropping so that the fused cloud respects fixed budgets for scene and robot points.
Geometric augmentations consist of random yaw rotations, isotropic scaling, and reflections; photometric augmentations apply auto-contrast, global color shifts, and per-point jitter to the RGB channels.
For evaluation, we ensure the pipeline to be fully deterministic and disable all augmentations.

\paragraph{Per-Point Input Features.}
Here we describe the per-point features produced as part of the data pipeline, prior to their consumption by the model.
Details are listed in Table~\ref{tab:features}.
Robot features stack positions, surface normals, a gripper scalar, and velocity and acceleration terms:
\[
\phi^{\text{robot}}_{t,j}
=
\bigl[
p^{\text{robot}}_{t,j},\,
c^{\text{robot}}_{j},\,
n^{\text{robot}}_{t,j},\,
\tilde{g}_t,\,
v^{\text{robot}}_{t,j},\,
a^{\text{robot}}_{t,j}
\bigr],
\]
where $p^{\text{robot}}_{t,j}$ and $n^{\text{robot}}_{t,j}$ are position and normal, $c^{\text{robot}}_{j}$ is a fixed color tag, $\tilde{g}_t$ is the normalized gripper openness, and $(v^{\text{robot}}_{t,j}, a^{\text{robot}}_{t,j})$ come from mid-point finite differences across the horizon with zero-velocity boundary conditions at the first and last timestep, i.e., we assume the robot is stationary at the boundaries of each model window.
Scene features are computed for only the first frame $t{=}0$ and combine positions, colors, estimated normals, gripper openness sequence, and distances to the nearest robot point:
\[
\phi^{\text{scene}}_{i}
=
\bigl[
x_{0,i},\,
c^{\text{scene}}_{0,i},\,
n^{\text{scene}}_{0,i},\,
g_{0{:}T-1},\,
d_{0{:}T-1,i}
\bigr],
\]
where $c^{\text{scene}}_{0,i}$ and $n^{\text{scene}}_{0,i}$ are the RGB color and normal at the first frame, $g_{0{:}T-1}\in\mathbb{R}^{T}$ is the sequence of gripper openness values over the context-plus-prediction horizon broadcast to every scene point, and $d_{t,i}$ is the distance from scene point $i$ to the closest robot point at timestep $t$,
\[
d_{t,i}
=
\min_{j}\,\bigl\|x_{0,i} - r_{t,j}\bigr\|_2,
\qquad
d_{0{:}T-1,i}\in\mathbb{R}^{T}.
\]
The distance field $d_{0{:}T-1,i}$ is obtained from nearest-neighbor queries between first-frame scene points and robot points at every timestep.

\paragraph{3D Scene Featurization with DINOv3.}
Prior to the point cloud backbone, \algo uses a 2D scene encoder based on DINOv3 ViT-L/16 by aggregating its multi-layer features.
To featurize the 3D scene points with the image-based encoder, the first-frame scene coordinates $x_{0,i}\in\mathbb{R}^3$ are projected into each chosen camera.
For camera $c$ with intrinsics $K_c$ and extrinsics $(R_c, t_c)$ we form $\tilde{u}_{c,i} = K_c (R_c x_{0,i} + t_c)$ and obtain the pixel coordinate as
\[
u_{c,i}
=
\bigl[\tilde{u}_{c,i}^{(1)}/\tilde{u}_{c,i}^{(3)},\,
\tilde{u}_{c,i}^{(2)}/\tilde{u}_{c,i}^{(3)}\bigr]^\top.
\]
The same intrinsics and extrinsics support a depth-consistency mask that compares the projected depth of each point with the given depth image, so only views whose discrepancy is below a few millimeters contribute features.

For each visible point-camera pair $(i,c)$, DINOv3 patch tokens are sampled at $u_{c,i}$ by bilinear interpolation on the patch-token grid, using a coordinate mapping that aligns token centers with pixel centers.
Let $f_{c,i}\in\mathbb{R}^{D_{\text{patch}}}$ denote the concatenated multi-layer patch feature for point $i$ in camera $c$, and let $m_{c,i}\in\{0,1\}$ indicate visibility and depth consistency.
Features are aggregated across cameras by averaging over the contributing views,
\[
f_i
=
\frac{1}{\max\bigl(1,\sum_c m_{c,i}\bigr)}
\sum_c m_{c,i}\,f_{c,i}.
\]
The averaged token is mapped to the backbone width (256 channels) by a learned projection and fused with a separately projected version of the raw scene features from Table~\ref{tab:features}; layer normalization is applied to each stream before concatenation, and a final linear layer produces the per-point embedding supplied to the dynamics backbone.
The 2D encoder is kept frozen during training and evaluation.

\paragraph{Visibility-Aware Supervision.}
For real-world domains, we restrict training on 3D point trajectories to correspondences that are both geometrically and photometrically reliable.
The annotation pipeline supplies per-point visibility (from 2D trackers~\cite{karaev2025cotracker3} on real data and from ground-truth simulator state on synthetic data) together with per-pixel depth-validity; both signals are propagated to the lifted 3D trajectories and stored as binary flags per scene point and timestep.
During training, we construct a per-timestep mask that selects scene points that are visible in the camera view and have valid depth support.
The weighted dynamics objective from Section~\ref{sec:method} is then evaluated only over this subset of correspondences (points filtered out receive zero loss weight), so that gradients are driven by non-occluded, depth-valid 3D flows.
For simulation domains, where trajectories and depth are noise-free and occlusions are explicitly modeled, all scene points contribute to the loss.

\paragraph{Training Configuration.}
We train the 1B-parameter version of \algo on both BEHAVIOR-1K and DROID, with configuration and PointTransformerV3 (PTv3) design summarized in Table~\ref{tab:train-config-1b} and Table~\ref{tab:ptv3-1b}, respectively.
For the main experiments in the paper, training configuration and PTv3 design are summarized in Table~\ref{tab:train-config} and Table~\ref{tab:ptv3-xl}.

\paragraph{Aleatoric Uncertainty on Simulation Data.}
When training on mixtures of real and simulated domains, directly learning per-point uncertainty everywhere can collapse the model because simulated trajectories are noise-free.
In the objective from Section~\ref{sec:method} the residual term for point $i$ at step $k$ is weighted as $w_{k,i}\,\rho_\delta(\mathbf{\hat P}_{t+k,i} - \mathbf{P}_{t+k,i})\,\mathrm{e}^{-s_{k,i}} + w_{k,i}\,s_{k,i}$.
For vanishing residuals (typical in simulation), minimizing the loss drives $s_{k,i}$ toward $\log \rho_\delta(\cdot)$.
Since $\rho_\delta(\cdot)$ approaches zero, the optimal $s_{k,i}$ becomes a large negative number, i.e., the predicted variance $\sigma^2_{k,i} = \exp(s_{k,i})$ collapses toward zero.
As a consequence, $\mathrm{e}^{-s_{k,i}}$ explodes, so any small numerical discrepancy in simulated residuals produces excessively large gradients that overwhelm the real-data contributions and destabilize joint training.
To stabilize training, we treat aleatoric variance on simulation domains as a constant: the uncertainty head is trained normally on real data, but for simulated domains its log-variance is replaced by a batch-wise constant that matches the average variance observed on real samples (or a small fixed value when only simulation is present).
This preserves heteroscedastic weighting where it is most useful (real, noisy supervision) while preventing the model from exploiting the uncertainty head to down-weight clean simulated gradients.

\begin{table*}[t]
    \centering
    \setlength{\tabcolsep}{4pt}
    \small

    \begin{tabularx}{0.8\linewidth}{l X}
        \toprule
        \textbf{Setting} & \textbf{Value} \\
        \midrule
        Optimizer & AdamW \\
        Learning rate & $1\times10^{-4}$ \\
        Epochs & 300 \\
        Weight decay & $10^{-2}$ \\
        Global batch size & 1920 sequences \\
        Gradient clipping & Global $\ell_2$ norm capped at $5$ \\
        Loss & Huber loss with $\delta=5.0$ with movement weighting and aleatoric uncertainty \\
        Prediction horizon & 10 steps \\
        Training GPUs & 128 NVIDIA H100 GPUs \\
        Training time & 20 days \\
        \bottomrule
    \end{tabularx}
    \caption{\textbf{Training Configuration for \algo-1B.}}
    \label{tab:train-config-1b}

    \vspace{0.75em}
    \begin{tabularx}{0.7\linewidth}{l X}
        \toprule
        \textbf{Component} & \textbf{Values} \\
        \midrule
        Grid size & $1.5$ cm \\
        Encoder depth & $(4,\ 4,\ 8,\ 8,\ 12,\ 12,\ 4)$ \\
        Encoder channels & $(256,\ 384,\ 384,\ 512,\ 512,\ 768,\ 1024)$ \\
        Encoder heads & $(8,\ 12,\ 12,\ 16,\ 16,\ 24,\ 32)$ \\
        Encoder stride & $(1,\ 2,\ 2,\ 2,\ 2,\ 2,\ 2)$ \\
        Encoder patch size & $(256,\ 256,\ 256,\ 256,\ 256,\ 256,\ 256)$ \\
        Decoder depth & $(4,\ 4,\ 4,\ 4,\ 4,\ 4)$ \\
        Decoder channels & $(256,\ 384,\ 384,\ 512,\ 512,\ 768)$ \\
        Decoder heads & $(8,\ 12,\ 12,\ 16,\ 16,\ 24)$ \\
        Decoder patch size & $(256,\ 256,\ 256,\ 256,\ 256,\ 256)$ \\
        \bottomrule
    \end{tabularx}
    \caption{\textbf{PointTransformerV3 (PTv3) Architecture for \algo-1B. Encoder and decoder configurations are ordered by stage.}}
    \label{tab:ptv3-1b}

    \vspace{0.75em}
    \begin{tabularx}{0.8\linewidth}{l X}
        \toprule
        \textbf{Setting} & \textbf{Value} \\
        \midrule
        Optimizer & AdamW \\
        Learning rate & $1\times10^{-4}$ \\
        Epochs & 200 \\
        Weight decay & $10^{-2}$ \\
        Global batch size & 176 sequences \\
        Gradient clipping & Global $\ell_2$ norm capped at $5$ \\
        Loss & Huber loss with $\delta=5.0$ with movement weighting and aleatoric uncertainty \\
        Prediction horizon & 10 steps \\
        Training GPUs & 8 NVIDIA H100 GPUs \\
        Training time & 7 days \\
        \bottomrule
    \end{tabularx}
    \caption{\textbf{Training Configuration for \algo-411M.}}
    \label{tab:train-config}

    \vspace{0.75em}
    \begin{tabularx}{0.7\linewidth}{l X}
        \toprule
        \textbf{Component} & \textbf{Values} \\
        \midrule
        Grid size & $1.5$ cm \\
        Encoder depth & $(4,\ 4,\ 4,\ 8,\ 8,\ 12,\ 4)$ \\
        Encoder channels & $(256,\ 256,\ 256,\ 384,\ 384,\ 512,\ 768)$ \\
        Encoder heads & $(4,\ 4,\ 4,\ 8,\ 8,\ 16,\ 24)$ \\
        Encoder stride & $(1,\ 2,\ 2,\ 2,\ 2,\ 2,\ 2)$ \\
        Encoder patch size & $(256,\ 256,\ 256,\ 256,\ 256,\ 256,\ 256)$ \\
        Decoder depth & $(2,\ 2,\ 2,\ 2,\ 2,\ 2)$ \\
        Decoder channels & $(256,\ 256,\ 256,\ 384,\ 384,\ 512)$ \\
        Decoder heads & $(4,\ 4,\ 4,\ 8,\ 8,\ 16)$ \\
        Decoder patch size & $(256,\ 256,\ 256,\ 256,\ 256,\ 256)$ \\
        \bottomrule
    \end{tabularx}
    \caption{\textbf{PointTransformerV3 (PTv3) Architecture for \algo-411M. Encoder and decoder configurations are ordered by stage.}}
    \label{tab:ptv3-xl}
\end{table*}

\appclearpage
\subsection{DROID Evaluation Protocol}
\label{app:data:eval}

Following the protocol in Section~\ref{sec:experiments}, we measure per-sequence losses and aggregate them into dataset-level summaries.
Alongside the overall per-point, per-timestep $\ell_2$ distance, we report the same metric separately on moved and static points, since movers form a minority of the points but dominate perceived quality.
We use these metrics directly for simulation data (BEHAVIOR-1K) since they are noiseless.
However, for real-world data (DROID), we further apply expert confidence filtering to obtain filtered metrics. Details are described below.

\paragraph{Expert Confidence Filtering.}
Although the mover-$\ell_2$ score highlights the behavior we care about most, imperfect real-world annotations mean that a noticeable fraction of mover points correspond to outliers or background clutter, because those points tend to have large movement magnitudes due to unstable depth estimation.
During training, the aleatoric uncertainty regularization down-weights those points, but at evaluation time different models produce their own uncertainty predictions, making comparisons challenging.
To obtain a model-agnostic notion of trustworthy ground-truth, we train an expert model only on the evaluation split with uncertainty predictions, convert the predicted variance into a per-point confidence in $[0,1]$, and threshold this per-timestep per-point confidence at the $0.8$ quantile over all points.
Points below this confidence are treated as low-confidence outliers.
We voxelize these low-confidence sets in world coordinates using the training grid size $g$ and cache the resulting voxel grids for each evaluation sample so that the same filtering masks can be reused across subsequent evaluation runs and model variants.
Note that the expert model is only used to compute the low-confidence voxel grids and does not share any training data or parameters with any evaluated models.

\paragraph{Filtered Evaluation.}
To evaluate a model, for each sample, we first reconstruct world-coordinate voxel indices of scene points and then determine whether each point lies inside a precomputed low-confidence voxel.
This yields a binary filter mask so that only high-confidence points at high-confidence timesteps (deemed by the shared expert model) contribute to the filtered metrics.

\paragraph{Mover/Static Splits.}
Let $\hat{P}_{t,i}$ and $P_{t,i}$ denote predicted and ground-truth 3D positions.
We compute per-point error $e_{t,i}=\lVert \hat{P}_{t,i}-P_{t,i}\rVert_2$ and report
\[
\ell_2 = \frac{1}{T}\sum_t \frac{1}{|V_t|}\sum_{i\in V_t} e_{t,i},
\]
where $V_t$ denotes valid points at timestep $t$.
Mover-$\ell_2$ and static-$\ell_2$ use the same definition but restrict $V_t$ to moved or static points identified from the ground-truth trajectories via a small displacement threshold.

\appclearpage
\subsection{Real-Robot Experiment Details}

Real-robot experiments use a 7-DoF Franka arm equipped with a 3D-printed fin ray gripper~\cite{diffusion_policy_2023,umi2024}.
The robot is mounted on a wheeled, non-motorized base for in-the-wild deployments.
Since \algo is trained on data containing Robotiq 2F-85 and Galexea R1 Pro grippers, the fin ray gripper geometry remains fully unseen by the model, illustrating cross-gripper geometry generalization.
Since the pipeline predicts 6-DoF end-effector poses, we run position control at $20$\,Hz: each predicted target pose is clipped to a predefined workspace, then linearly interpolated from the current pose with steps of $5$\,mm in translation and $1^\circ$ in rotation.
For every interpolated pose, inverse kinematics (PyBullet solver) produces target joint positions that are tracked with the Deoxys joint-impedance controller~\cite{zhu2022viola}.
We use one RealSense D435 mounted on the left shoulder of the robot to capture RGB and the stereo IR images.
The stereo IR images are used to estimate the metric depth using FoundationStereo~\citep{foundationstereo}, given known baseline and camera intrinsics.

\subsubsection{Model-Based Planning}
In this work, we use a single pre-trained \algo as the dynamics model.
The model is pre-trained jointly on both real-world and simulated data.
We use a sampling-based model-predictive path integral (MPPI) controller that samples action sequences around a zero-initialized nominal using cubic splines with $n_{\text{knots}}{=}4$ and degree $3$.
Noise scales are scheduled between $\sigma_{\min}{=}0.05$ and $\sigma_{\max}{=}0.50$ (in normalized action units).
Each refinement iteration draws 256 samples; importance weights use temperature $\beta{=}0.05$, and the nominal is updated with an exponential moving average (EMA $=0.9$).
We perform planning for $30$ steps into the future, and the horizon is chunked to match the prediction window of the dynamics model.
We perform $20$ refinement iterations.
The planning time is typically around a few seconds depending on task complexity and specific model size variant used.
While we do not perform replanning in this work, replanning can be done at a real-time frequency by warm-starting from the previous nominal trajectory.

\subsubsection{Task Specification}
We specify tasks through a GUI tool that allows users to select object masks using SAM2~\cite{ravi2024sam2} and specify target positions in the world frame.
We find this simple objective as a unified interface for specifying diverse real-world tasks including rigid pushing, deformable manipulation, articulated-object interaction, and tool use.
Following common practices in reward design~\cite{zakka2025mujoco}, we add a mild end-effector proximity term to encourage exploration in the object’s neighborhood without prescribing a particular contact pattern.
For deformable and tool-use tasks we begin from a pre-grasped configuration so that subsequent motion primarily probes deformable dynamics and object-object contacts.
All tasks share same control regularization comprising SE(3) path-length penalties and IK-based reachability residual.

\subsubsection{Evaluation Protocol}
We conduct evaluations on the following tasks: rigid pushing (tissue box, book), deformable manipulation (scarf fold, pillow place), articulated manipulation (microwave open, drawer close), and tool use (duster sweep, broom sweep).
Each task is evaluated with ten randomly sampled initial configurations.
The configurations are sampled prior to evaluation and verified to be kinematically feasible for the robot.
For each trial, a human operator restores the scene to the designated configuration and triggers execution.
We consider the trial successful if the task objective is met.
Otherwise we declare failure.
If the optimization produces a solution that is considered unsafe for execution, the trial is considered failure too.
The success rates are reported in the main paper.

\subsubsection{Effect of Training Mixture}
Beyond the quantitative success rates for real-world deployment, we observe interesting qualitative traits when using different variants of \algo pre-trained on different data mixtures.
We empirically observe that models trained only on real data tend to be conservative: a common failure mode is for scene points to remain static even when the robot establishes contact, which we attribute to heavy regularization coping with annotation noise.
On the other hand, models trained only on simulation data excel on rigid objects but frequently mis-segment cluttered real scenes implicitly, causing background points to move together with the target.
Models trained on both real and simulated domains yield the most balanced behavior in practice, combining realistic contact handling with the ability to generalize to novel real-world scenes.
A systematic study of how training-mixture design shapes deployment-time behavior, e.g., by varying real/sim proportions or task/domain coverage under controlled conditions, remains an important direction for future work.

\appclearpage
\subsection{Additional 3D Annotation Examples}
\label{app:annotation-examples}
Interactive visualizations available at \href{https://point-world.github.io}{project website}.

\begin{figure*}[b]
    \centering
    \includegraphics[width=\linewidth]{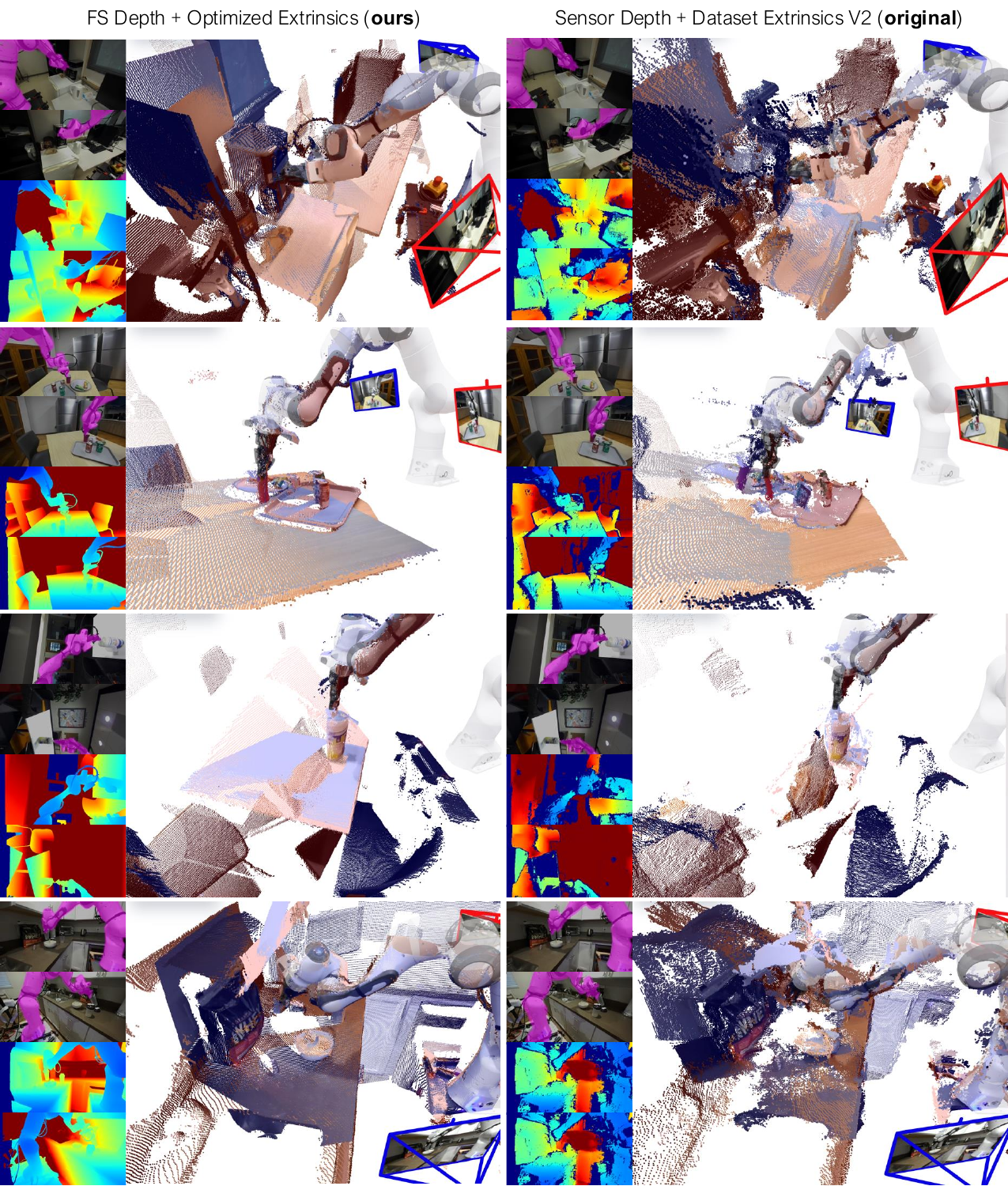}
    \caption{\textbf{DROID 3D annotations}, including robot-overlaid RGBs, depths, point clouds, and comparisons to original dataset.}
    \label{fig:annotation-app1}
\end{figure*}
\appclearpage

\begin{figure*}[b]
    \centering
    \includegraphics[width=\linewidth]{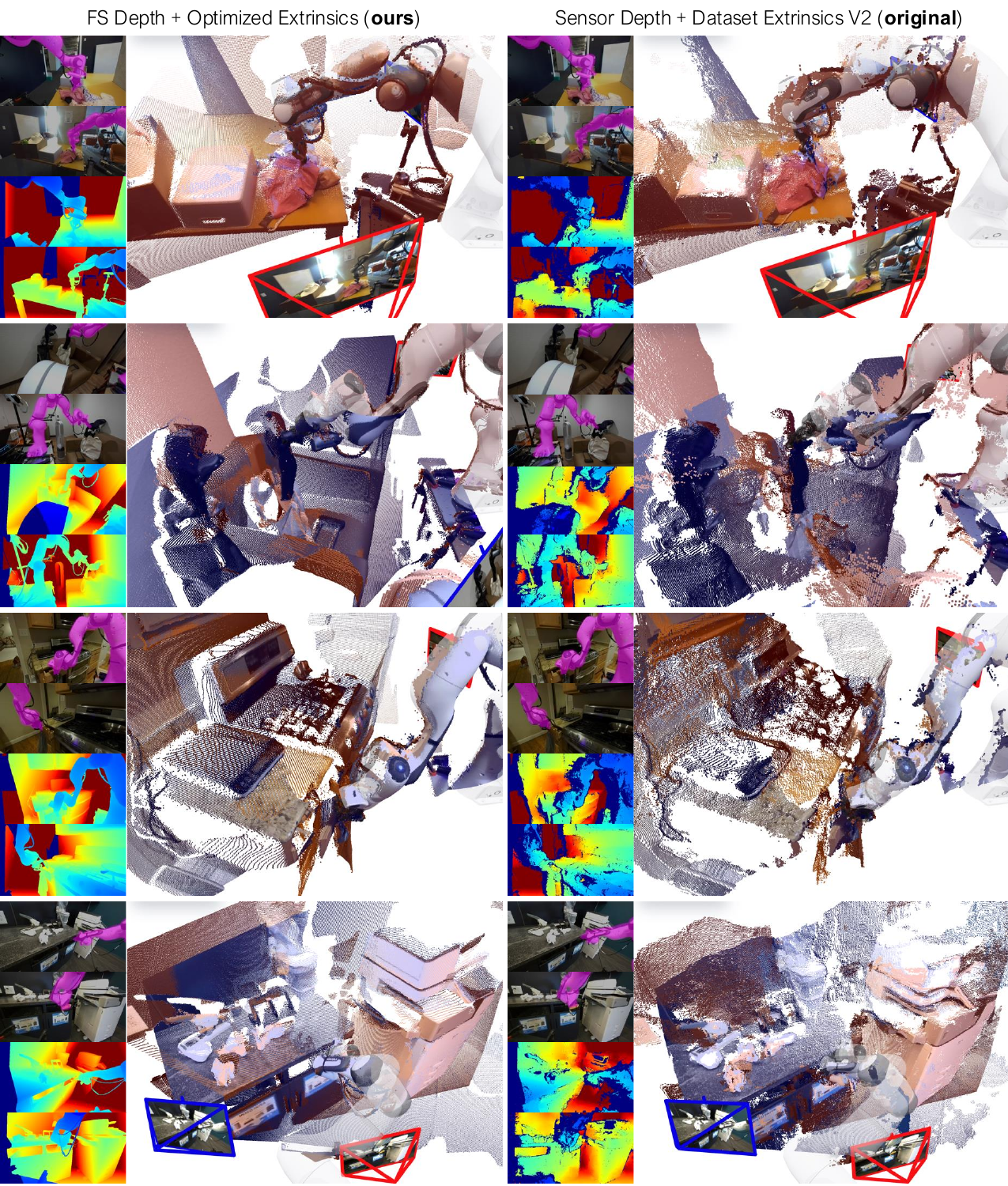}
    \caption{\textbf{DROID 3D annotations}, including robot-overlaid RGBs, depths, point clouds, and comparisons to original dataset.}
    \label{fig:annotation-app2}
\end{figure*}
\appclearpage

\subsection{Additional Model Rollouts}
\label{app:rollouts}
Interactive visualizations available at \href{https://point-world.github.io}{project website}.

\begin{figure*}[b]
    \centering
    \includegraphics[width=\linewidth]{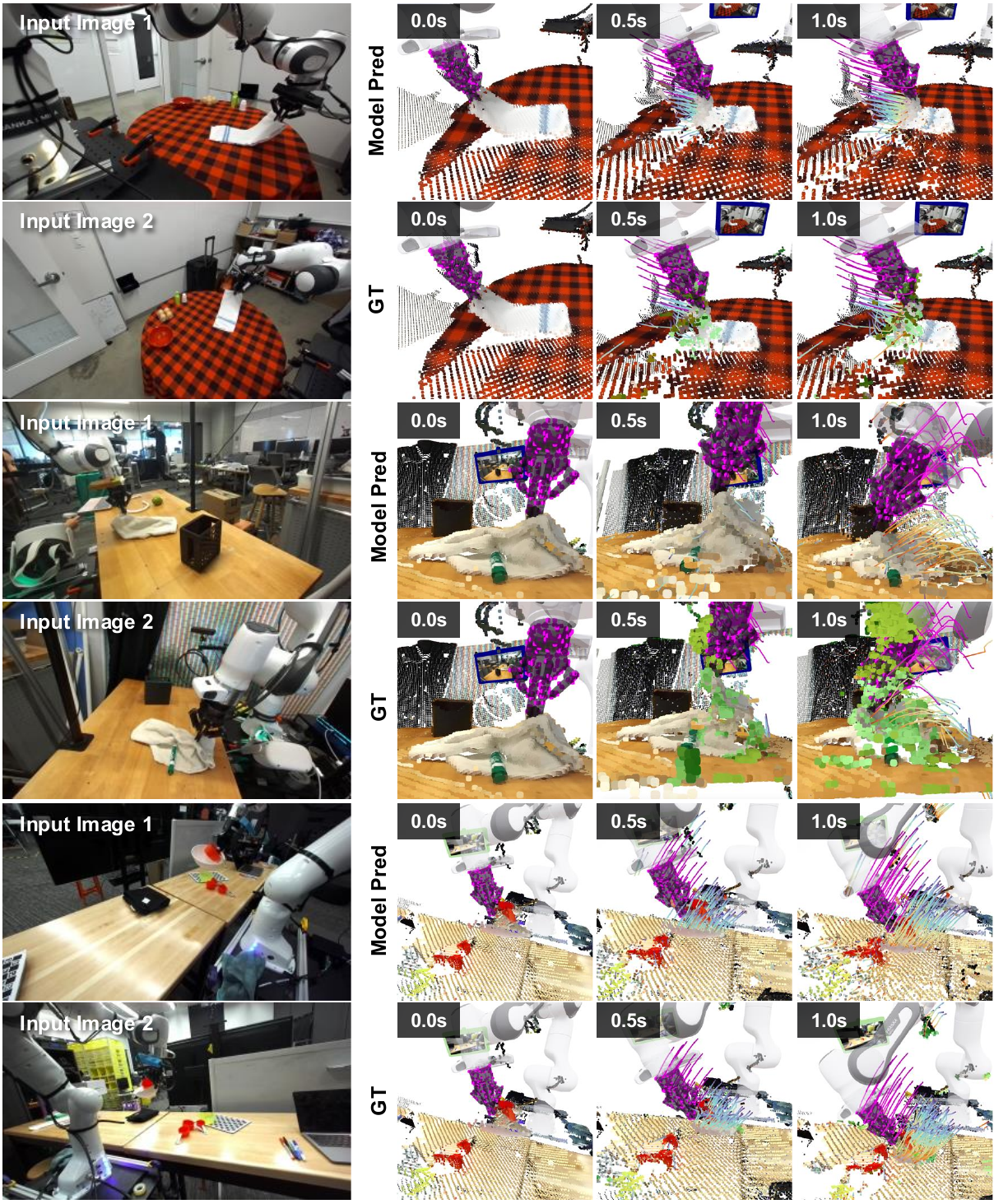}
    \vspace{-1.5em}
    \caption{\textbf{DROID Unseen Rollouts}, including deformable manipulation, robot-object interactions, and object-object interactions.}
    \vspace{-1.5em}
    \label{fig:rollouts-app1}
\end{figure*}

\begin{figure*}[b]
    \centering
    \includegraphics[width=\linewidth]{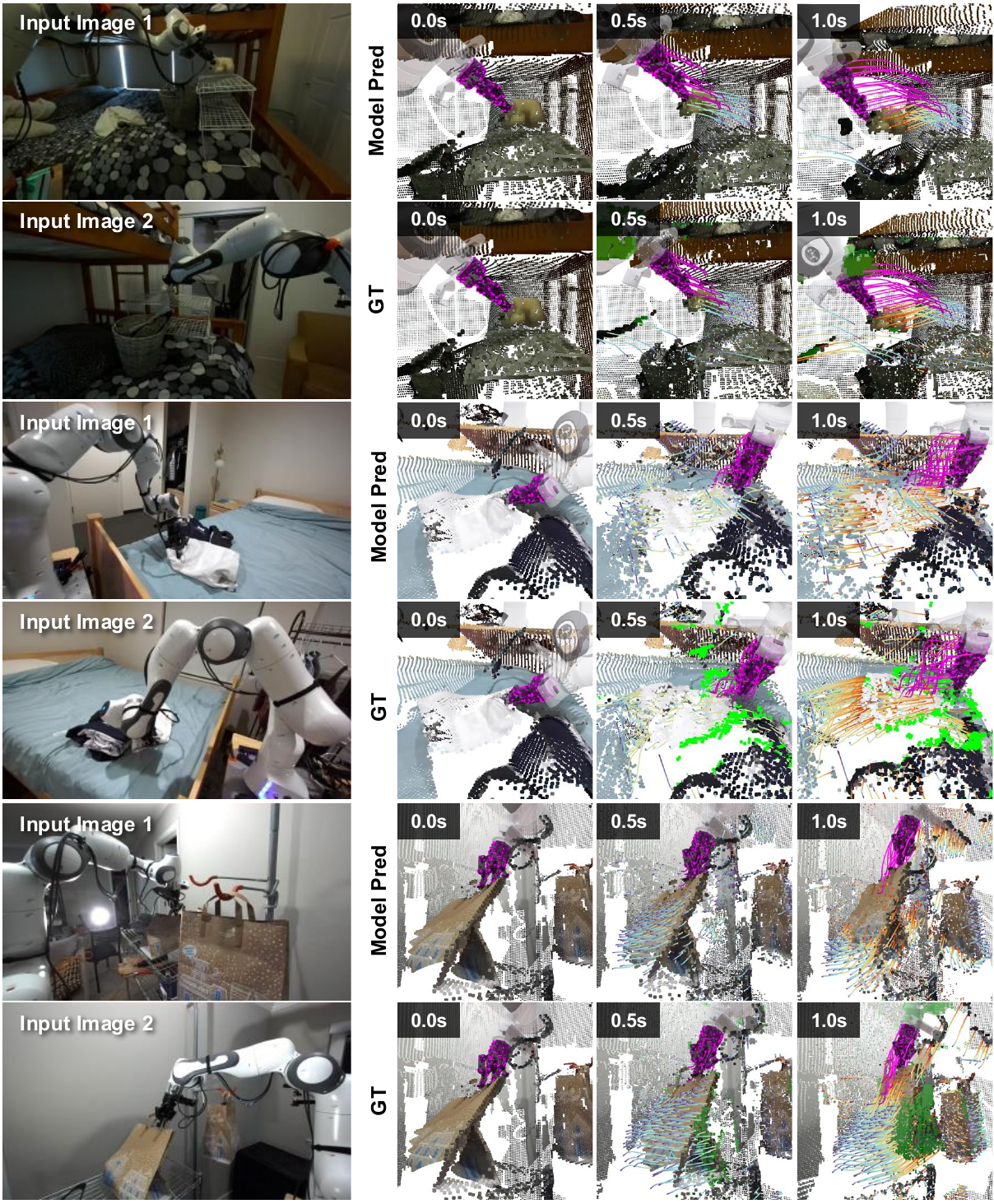}
    \caption{\textbf{DROID Unseen Rollouts}, including deformable manipulation, and grasping behaviors.}
    \label{fig:rollouts-app2}
\end{figure*}

\begin{figure*}[b]
    \centering
    \includegraphics[width=\linewidth]{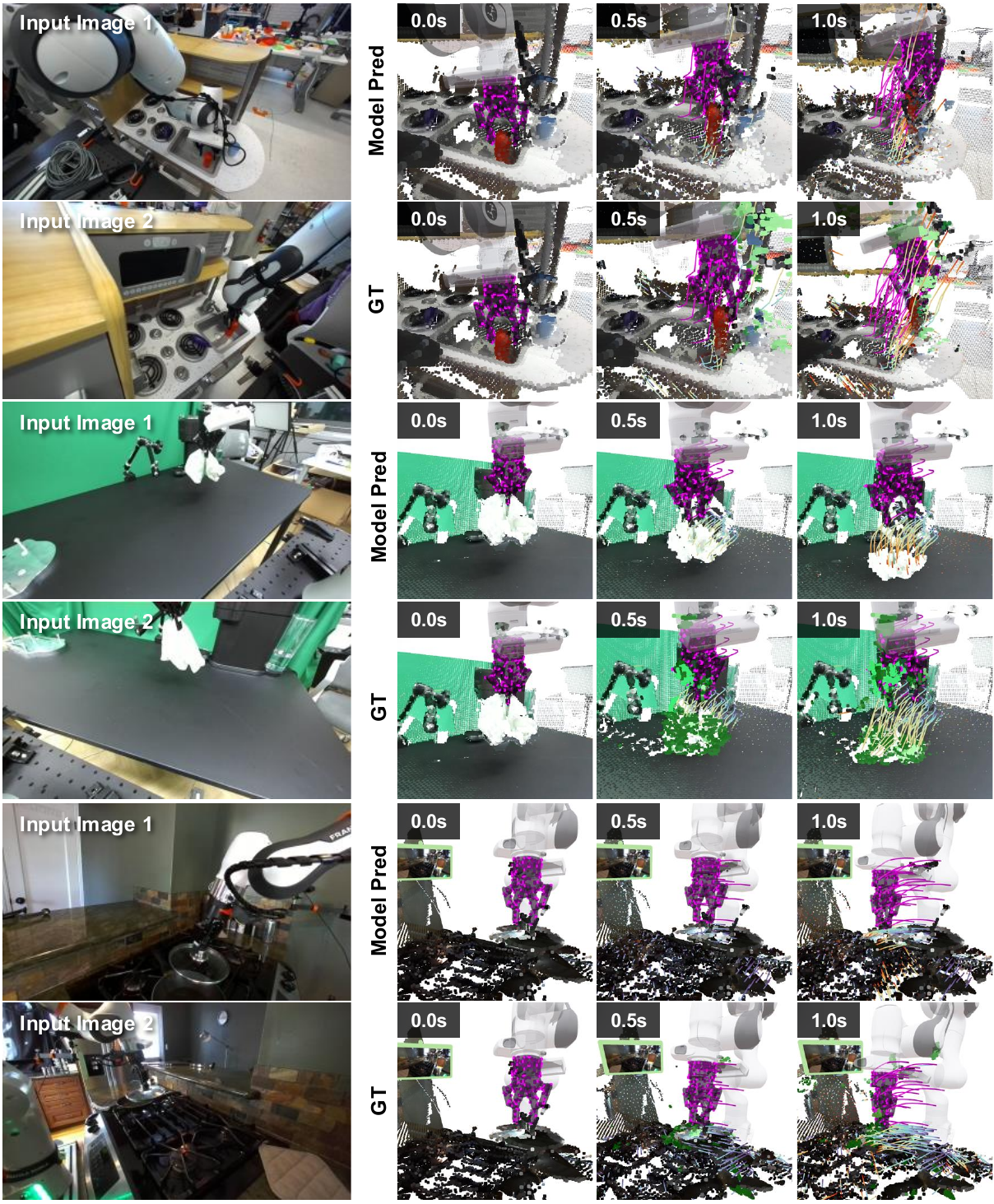}
    \caption{\textbf{DROID Unseen Rollouts}, including grasping behaviors, gravity effects, and glass objects.}
    \label{fig:rollouts-app3}
\end{figure*}

\begin{figure*}[b]
    \centering
    \includegraphics[width=\linewidth]{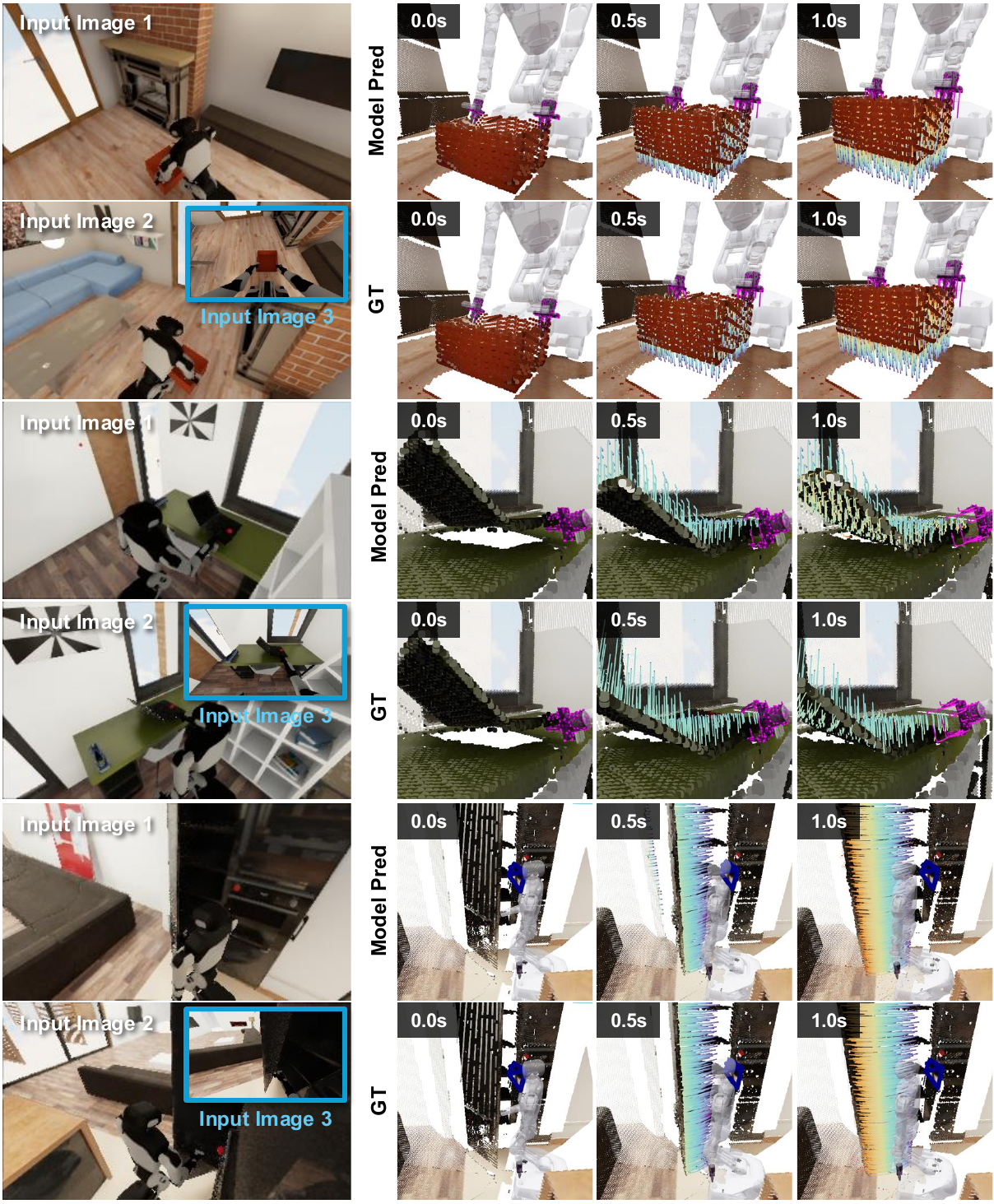}
    \caption{\textbf{BEHAVIOR-1K Unseen Rollouts}, including constrained bimanual lifting, gravity effects (dropped laptop), object-object interactions (laptop v.s. table), and articulated manipulation (fridge).}
    \label{fig:rollouts-app4}
\end{figure*}

\begin{figure*}[b]
    \centering
    \includegraphics[width=\linewidth]{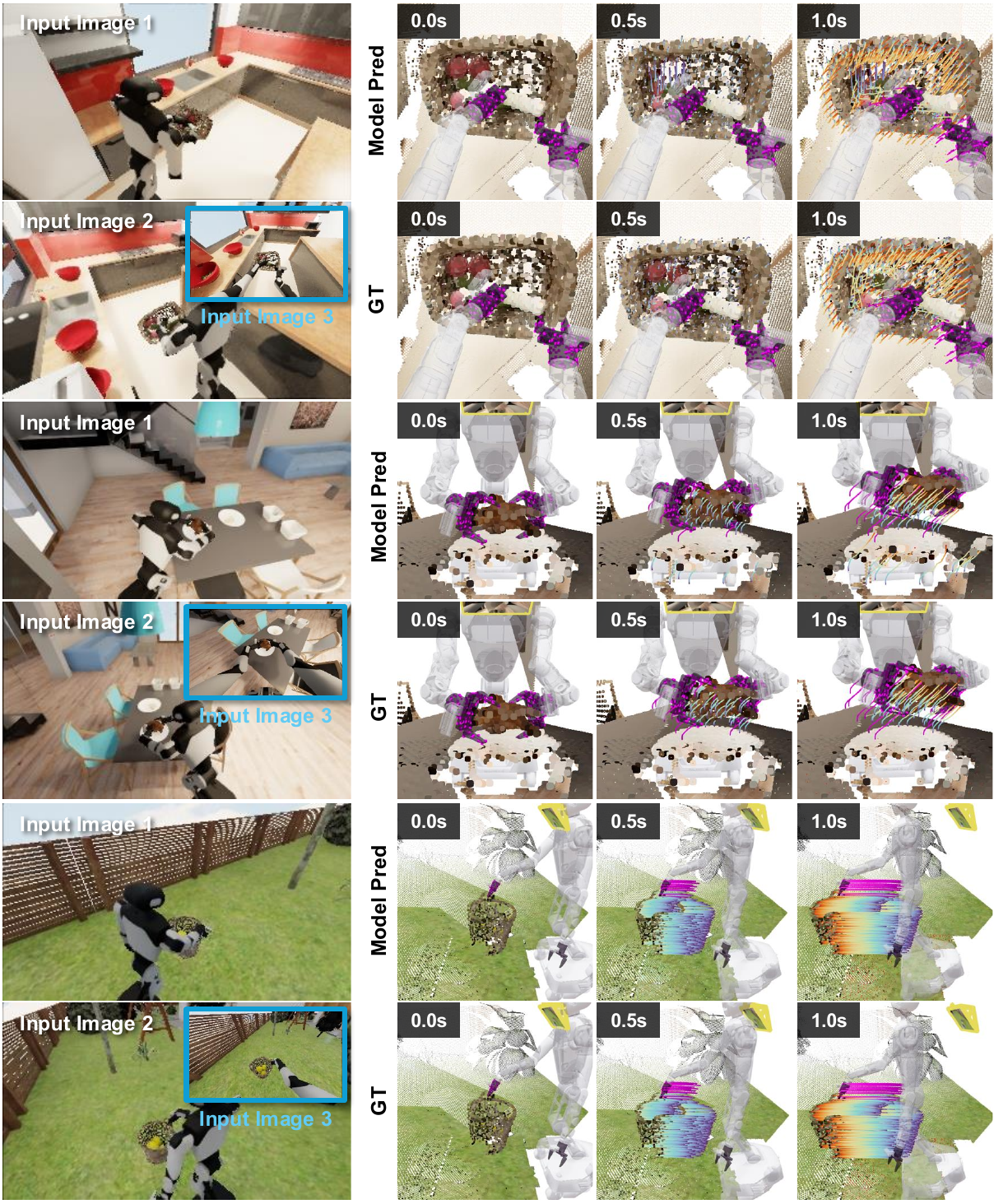}
    \caption{\textbf{BEHAVIOR-1K Unseen Rollouts}, including object-object interactions (within basket), gravity effects (within basket), and whole-body behaviors.}
    \label{fig:rollouts-app5}
\end{figure*}

\begin{figure*}[b]
    \centering
    \includegraphics[width=\linewidth]{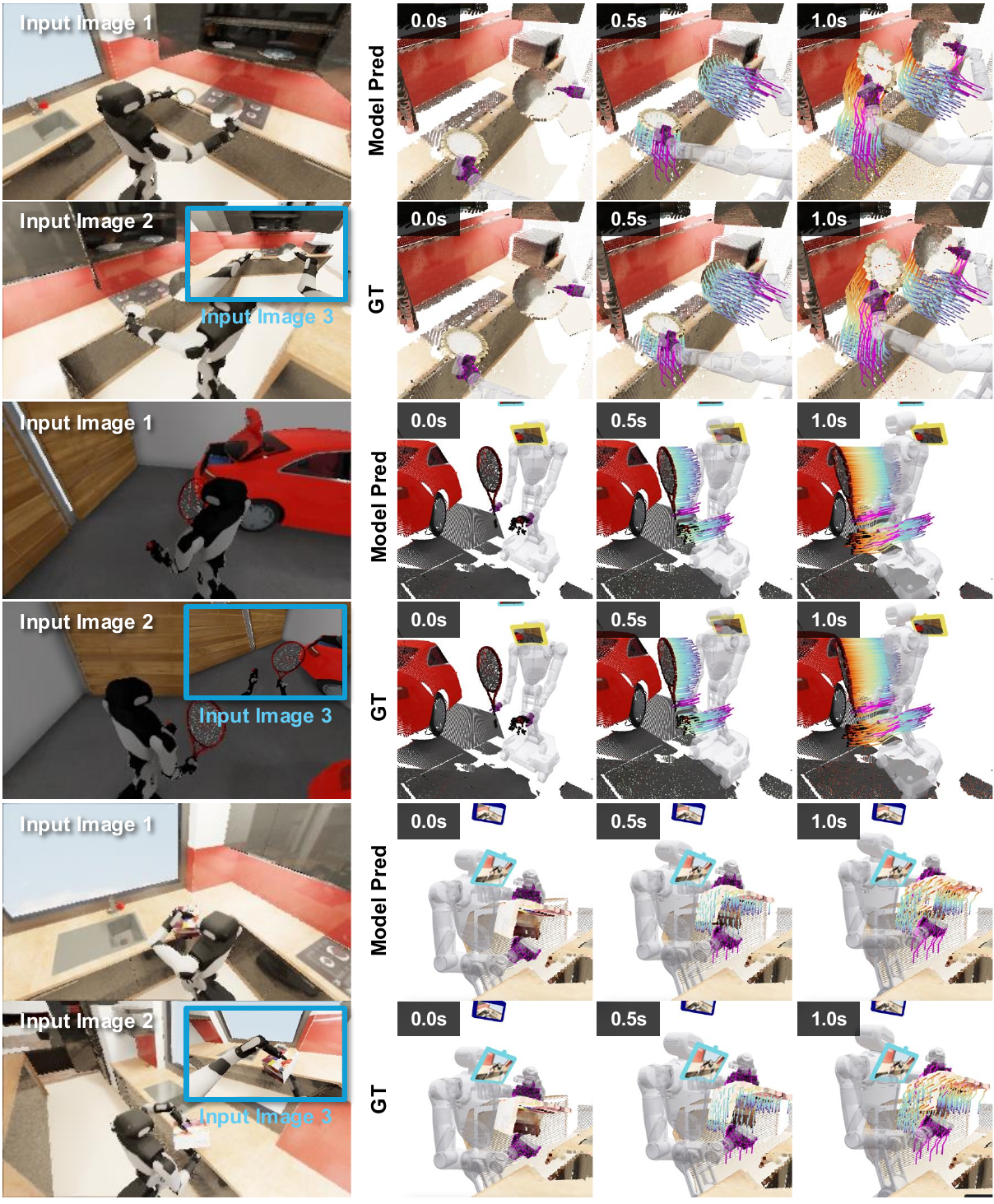}
    \caption{\textbf{BEHAVIOR-1K Unseen Rollouts}, including bimanual manipulation, whole-body behaviors, and implicit shape completion.}
    \label{fig:rollouts-app6}
\end{figure*}

\end{document}